\documentclass[journal]{IEEEtran}
\usepackage{amsmath,amsfonts,amssymb}
\usepackage{algorithmic}
\usepackage{array}
\usepackage[caption=false,font=normalsize,labelfont=sf,textfont=sf]{subfig}
\usepackage{textcomp}
\usepackage{stfloats}
\usepackage{url}
\usepackage{verbatim}
\usepackage{graphicx}
\usepackage{cite}
\usepackage{xcolor}
\usepackage{multirow}
\usepackage{balance}
\usepackage{placeins}
\usepackage[linesnumbered,ruled,vlined]{algorithm2e}
\usepackage{tikz}
\usepackage{booktabs}
\usepackage{pgfplots}
\usepackage{booktabs}
\usepackage{tabularx}
\usepackage{makecell}
\usepackage{array}
\usetikzlibrary{arrows.meta,positioning,fit,calc,backgrounds}
\usepgfplotslibrary{groupplots}
\pgfplotsset{compat=1.18}
\newcolumntype{L}[1]{>{\raggedright\arraybackslash}p{#1}}

\providecommand{\best}[1]{\textbf{#1}}

\begin{document}

\title{TypeBandit: Type-Level Context Allocation and Reweighting for Effective Attribute Completion in Heterogeneous Graph Neural Networks}

\author{Ta-Yang Wang, Rajgopal Kannan, Viktor Prasanna
\thanks{Ta-Yang Wang and Viktor Prasanna are with University of Southern California, Los Angeles, USA. (e-mail: {tayangwa, prasanna}@usc.edu)}
\thanks{Rajgopal Kannan is with DEVCOM Army Research Office, Los Angeles, USA. (e-mail: rajgopal.kannan.civ@army.mil)}
\thanks{\textbf{Distribution Statement A:} Approved for public release. Distribution is unlimited.}
}

\maketitle

\markboth{}{}

\begin{abstract}
Heterogeneous graphs are widely used to model multi-relational systems, but missing node attributes remain a major bottleneck for downstream learning. In this paper, we identify and formalize \emph{type-dependent information asymmetry}: the phenomenon that different node types provide substantially different levels of useful signal for attribute completion. Motivated by this observation, we propose TypeBandit, a lightweight, model-agnostic methodology for heterogeneous attribute completion. TypeBandit combines topology-aware initialization, type-level bandit sampling, and joint representation learning. It allocates a finite global sampling budget across node types, samples representative nodes within each type, and uses the resulting sampled type summaries as shared contextual signals during representation construction. By operating at the type level rather than over each target node's local neighborhood, TypeBandit keeps the adaptive state compact and computationally practical for large heterogeneous graphs.

A key advantage of TypeBandit is architectural flexibility. Rather than requiring a new heterogeneous graph neural network architecture, TypeBandit acts as a type-aware front end that can be integrated with representative heterogeneous GNN backbones, including R-GCN, HetGNN, HGT, and SimpleHGN. We further introduce a hybrid pretraining scheme that combines structural degree priors with feature propagation, yielding a more reliable initializer than degree-only pretraining. Under a fixed-split protocol on DBLP, IMDB, and ACM, TypeBandit provides dataset-dependent but practically meaningful gains, with the clearest improvements on DBLP, more modest gains on ACM, and initialization-sensitive behavior on IMDB. Additional ablation, stability, efficiency, semantic-propagation, and sampled OGBN-MAG experiments further support TypeBandit as a practical strategy for heterogeneous attribute completion when type-specific information is unevenly distributed and sampling resources are limited.
\end{abstract}


\section{Introduction}
\IEEEPARstart{H}{eterogeneous} graphs are extensively used to model complex systems with multiple node and edge types, including academic networks, recommendation systems, and multimedia knowledge graphs~\cite{wang2022efficient,wang2022survey,salamat2021heterographrec,li2022sybilflyover,wang2022throughput}. Their expressive power comes from the ability to encode different semantic roles and interaction patterns in a single graph. Graph Neural Networks (GNNs) have become a standard tool for representation learning on such data~\cite{wang2019heterogeneous2,zhang2019heterogeneous,hu2020heterogeneous,chen2019activehne,yan2024fast,wang2023enabling,bengre2021learning,li2024gnnri,duong2022hatenet}. However, in many practical heterogeneous graphs, only a subset of node types comes with observed attributes, while the remaining types must be inferred from structure and cross-type interactions. This attribute incompleteness can substantially degrade downstream learning.

Existing heterogeneous graph learning methods typically address missing attributes through mean aggregation, static feature propagation, or fixed meta-path heuristics~\cite{hu2020heterogeneous,fu2020magnn,jin2021heterogeneous}. These strategies can be useful, but they often treat all neighbor types as equally informative. In practice, this assumption is too coarse. The value of a neighbor for attribute completion depends strongly on its type and on the graph domain, i.e., the dataset-specific relational semantics. This leads to a central challenge that motivates our work: \textit{type-dependent information asymmetry}. Some types provide highly relevant semantic cues, whereas others contribute only weak or noisy signals.

We address this problem by viewing type-level resource allocation and sampling as an adaptive decision problem. In TypeBandit, each node type corresponds to an arm in a type-level bandit policy, and the learned policy weights determine how a finite global sampling budget is distributed across type pools. Within each type, the model samples representative nodes, averages them into a sampled type context, and fuses that context back into the latent representation before the heterogeneous backbone. This design is intentionally global rather than node-local: instead of running a separate bandit over every target node's heterogeneous neighborhood, which would be computationally prohibitive on large heterogeneous graphs, TypeBandit uses a compact type-level policy to decide which types deserve additional sampling effort. The resulting methodology combines four ingredients: (1) topology-aware initialization for nodes with missing attributes; (2) latent feature projection into a shared representation space; (3) a type-level sampling policy updated through reward-modulated multiplicative reweighting; and (4) a joint completion-and-prediction objective built on top of a heterogeneous backbone. In the implementation studied here, the initialization stage supports degree-only, feature-propagation, and hybrid pretraining, with the hybrid variant providing the most reliable overall behavior.

The main contributions of this paper are summarized as follows:
\begin{itemize}
\item We formulate type-dependent information asymmetry as a practical bottleneck in heterogeneous graph attribute completion and propose TypeBandit as a lightweight front end that combines topology-aware initialization, type-level budget allocation, and bandit-based sampling.
\item We introduce a hybrid pretraining scheme that combines structural degree priors with feature propagation, yielding a more robust and reliable initializer for missing-feature node types than degree-only pretraining.
\item We demonstrate that TypeBandit can be integrated with representative heterogeneous GNN backbones, including R-GCN, HetGNN, HGT, and SimpleHGN, without requiring a new backbone architecture.
\item We conduct a comprehensive empirical study under fixed-split protocols on DBLP, IMDB, and ACM, together with ablation, stability, efficiency, semantic-propagation, and sampled OGBN-MAG experiments to evaluate robustness and scalability.
\end{itemize}

Together, these results position TypeBandit as a plug-in front end for type-aware attribute completion and sampling-budget allocation. Its role is to complement existing heterogeneous GNN backbones by improving the construction of missing-type representations, rather than to replace the backbone architecture itself.

The rest of this paper is organized as follows:
Section~\ref{sec:ps} introduces the heterogeneous-graph setting and notation.
Section~\ref{sec:pd} formalizes the problem of adaptive heterogeneous attribute completion.
Section~\ref{sec:work} reviews the related literature.
Section~\ref{sec:method} presents the TypeBandit methodology.
Section~\ref{sec:exp} reports the experimental results,
and Section~\ref{sec:con} concludes the paper.

\section{Background}
\label{sec:ps}

We first introduce the heterogeneous graph setting, the attribute-missing problem, and the type-level adaptive sampling perspective used by TypeBandit. The main notations used throughout the paper are summarized in Table~\ref{table:notation}. Throughout the paper, calligraphic letters denote sets and bold capital letters denote matrices.

\begin{table}[t]
\caption{Notation Summary}
\label{table:notation}
\centering
\resizebox{\columnwidth}{!}{
\begin{tabular}{ll}
\hline
Notation & Description \\
\hline
$\mathcal{G}$ & Heterogeneous graph \\
$\mathcal{V}$ & Set of nodes \\
$\mathcal{E}$ & Set of edges \\
$\mathcal{O}_{\mathcal{V}}$ & Set of node types \\
$\mathcal{R}_{\mathcal{E}}$ & Set of edge types / relation types \\
$\varphi$ & Node-type mapping function, $\varphi:\mathcal{V}\rightarrow\mathcal{O}_{\mathcal{V}}$ \\
$\phi$ & Edge-type mapping function, $\phi:\mathcal{E}\rightarrow\mathcal{R}_{\mathcal{E}}$ \\
$\mathcal{V}_{o}$ & Set of nodes with type $o$ \\
$N(v)$ & Neighborhood of node $v$ \\
$N_o(v)$ & Typed neighborhood of $v$ containing neighbors of type $o$ \\
$N_r(v)$ & Relation-specific neighborhood of $v$ under relation type $r$ \\
\hline
$\mathcal{O}_{\mathcal{V}}^{+}$ & Set of node types with observed attributes \\
$\mathcal{O}_{\mathcal{V}}^{-}$ & Set of attribute-missing node types \\
$\mathcal{V}^{+}$ & Set of nodes whose types belong to $\mathcal{O}_{\mathcal{V}}^{+}$ \\
$\mathcal{V}^{-}$ & Set of nodes whose types belong to $\mathcal{O}_{\mathcal{V}}^{-}$ \\
$\mathbf{X}_{o}$ & Observed feature matrix of attributed node type $o$ \\
$d_o$ & Raw feature dimension of node type $o$ \\
$\mathbf{Z}^{(0)}_{o}$ & Completed or initialized latent feature matrix of node type $o$ \\
$d$ & Shared latent feature dimension \\
\hline
$\mathbf{h}^{(\ell)}_{v}$ & Hidden representation of node $v$ at layer $\ell$ \\
$\mathbf{H}$ & Final learned node embeddings \\
$\hat{\mathbf{Y}}$ & Predicted labels for the downstream task \\
$\mathcal{L}_{\mathrm{prediction}}$ & Downstream prediction loss \\
$\mathcal{L}_{\mathrm{completion}}$ & Attribute-completion or reconstruction-related loss \\
$\mathcal{L}_{\mathrm{total}}$ & Overall training objective \\
\hline
$\mathcal{A}$ & Type-level action space in the bandit policy \\
$K$ & Number of candidate node types, $K=|\mathcal{A}|$ \\
$w_t(o)$ & Bandit weight of node type $o$ at training step $t$ \\
$p_t(o)$ & Sampling probability of node type $o$ at training step $t$ \\
$p_{\min}$ & Minimum exploration probability for each type \\
$\eta$ & Bandit learning rate \\
$r_t(o)$ & Observed or estimated reward signal for node type $o$ \\
\hline
\end{tabular}
}
\end{table}

\subsection{Heterogeneous Graphs}

A heterogeneous graph is denoted by
\[
\mathcal{G}=(\mathcal{V},\mathcal{E},\mathcal{O}_{\mathcal{V}},\mathcal{R}_{\mathcal{E}}),
\]
where $\mathcal{V}$ is the set of nodes, $\mathcal{E}$ is the set of edges, $\mathcal{O}_{\mathcal{V}}$ is the set of node types, and $\mathcal{R}_{\mathcal{E}}$ is the set of edge types. Each node $v \in \mathcal{V}$ is associated with a type through a mapping function
\[
\varphi:\mathcal{V}\rightarrow\mathcal{O}_{\mathcal{V}},
\]
and each edge $e \in \mathcal{E}$ is associated with a relation type through
\[
\phi:\mathcal{E}\rightarrow\mathcal{R}_{\mathcal{E}}.
\]
For a node type $o \in \mathcal{O}_{\mathcal{V}}$, we denote by
\[
\mathcal{V}_{o}=\{v\in\mathcal{V}\mid \varphi(v)=o\}
\]
the set of nodes of that type. The typed neighborhood of a node $v$ with respect to node type $o$ is denoted by
\[
N_{o}(v)=\{u\in\mathcal{V}\mid (u,v)\in\mathcal{E},\ \varphi(u)=o\}.
\]
Similarly, for a relation type $r\in\mathcal{R}_{\mathcal{E}}$, we denote the relation-specific neighborhood of $v$ by
\[
N_r(v)=\{u\in\mathcal{V}\mid (u,v)\in\mathcal{E},\ \phi((u,v))=r\},
\]
when the backbone explicitly models edge types.

Heterogeneous graphs are more expressive than homogeneous graphs because they encode multiple semantic roles and relation patterns in a single graph. For example, an academic network may contain authors, papers, venues, and terms; a recommendation graph may contain users, items, categories, and interactions; and a multimedia knowledge graph may contain entities, concepts, images, and textual attributes. This semantic richness is precisely what makes heterogeneous graphs useful, but it also makes representation learning more sensitive to how information is propagated across types.

\subsection{Attribute-Missing Node Types}

In many heterogeneous graph benchmarks and applications, only a subset of node types is associated with observed attributes. Let
\[
\mathcal{O}_{\mathcal{V}}^{+}\subseteq \mathcal{O}_{\mathcal{V}}
\]
denote the set of attributed node types, and let
\[
\mathcal{O}_{\mathcal{V}}^{-}=\mathcal{O}_{\mathcal{V}}\setminus \mathcal{O}_{\mathcal{V}}^{+}
\]
denote the set of attribute-missing node types. Accordingly, the node set can be partitioned into
\[
\mathcal{V}^{+}=\{v\in\mathcal{V}\mid \varphi(v)\in\mathcal{O}_{\mathcal{V}}^{+}\},
\]
and
\[
\mathcal{V}^{-}=\{v\in\mathcal{V}\mid \varphi(v)\in\mathcal{O}_{\mathcal{V}}^{-}\}.
\]
For each attributed type $o\in\mathcal{O}_{\mathcal{V}}^{+}$, the observed feature matrix is denoted by
\[
\mathbf{X}_{o}\in\mathbb{R}^{|\mathcal{V}_{o}|\times d_o},
\]
where $d_o$ is the raw feature dimension of type $o$. For types in $\mathcal{O}_{\mathcal{V}}^{-}$, raw features are unavailable. The objective is therefore not necessarily to reconstruct the original raw attributes, which may not exist or may not be observed, but to infer effective latent representations that can support downstream learning.

We denote the initialized or completed latent feature matrix of type $o$ by
\[
\mathbf{Z}^{(0)}_{o}\in\mathbb{R}^{|\mathcal{V}_{o}|\times d},
\]
where $d$ is the shared latent dimension used by the heterogeneous encoder. For attributed node types, $\mathbf{Z}^{(0)}_{o}$ can be obtained by projecting observed features into the shared latent space. For attribute-missing node types, $\mathbf{Z}^{(0)}_{o}$ must be inferred from structural information, propagated features, or both. The quality of this initialization can have a substantial effect on the subsequent message-passing process.

\subsection{Heterogeneous Graph Neural Networks}

Heterogeneous Graph Neural Networks extend standard message passing by incorporating node and relation types into neighborhood aggregation. A generic heterogeneous message-passing layer can be written as
\[
\mathbf{h}_{v}^{(\ell+1)}
=
\sigma\left(
\mathbf{W}_{\varphi(v)}^{(\ell)}\mathbf{h}_{v}^{(\ell)}
+
\sum_{r\in\mathcal{R}_{\mathcal{E}}}
\sum_{u\in N_{r}(v)}
\alpha_{uvr}^{(\ell)}
\mathbf{W}_{r}^{(\ell)}\mathbf{h}_{u}^{(\ell)}
\right),
\]
where $\mathbf{h}_{v}^{(\ell)}$ is the representation of node $v$ at layer $\ell$, $N_r(v)$ is the relation-specific neighborhood of $v$, $\mathbf{W}_{r}^{(\ell)}$ is a relation-specific transformation, $\alpha_{uvr}^{(\ell)}$ is an aggregation weight, and $\sigma(\cdot)$ is a nonlinear activation function. Different heterogeneous backbones instantiate this template differently. For example, relation-aware GNNs use relation-specific transformations, attention-based models learn edge- or type-level attention weights, and simplified heterogeneous models may rely on precomputed semantic propagation.

When node attributes are incomplete, the effectiveness of heterogeneous message passing depends on two coupled factors. First, missing-feature node types require meaningful initial latent representations before they can participate effectively in message passing. Second, not all neighboring types contribute equally useful information. Some types may provide strong semantic cues for attribute completion, while others may introduce weak, redundant, or noisy signals. Therefore, heterogeneous attribute completion is not only an initialization problem, but also a type-aware information selection problem.

\subsection{Type-Level Adaptive Bandit Sampling}

A common strategy in graph representation learning is to sample or aggregate from neighborhoods according to a fixed rule, such as uniform sampling, mean aggregation, static feature propagation, or manually specified meta-paths. These strategies are simple and often effective, but they implicitly assume that the contribution of each neighbor type is either uniform or predetermined. This assumption can be too restrictive in heterogeneous graphs, where the usefulness of a neighbor depends heavily on its node type, relation semantics, and the downstream task.

TypeBandit adopts a type-level adaptive bandit-sampling perspective. Instead of treating individual nodes as separate actions, TypeBandit treats node types as the decision units. Let
\[
\mathcal{A}=\mathcal{O}_{\mathcal{V}}
\]
or a task-specific subset of node types denote the action space. At training step $t$, the sampler maintains a probability distribution
\[
\mathbf{p}_{t}\in\Delta^{|\mathcal{A}|-1}
\]
over node types. This distribution does not define a separate bandit decision for every target node and every local neighborhood. Instead, it allocates a finite global sampling budget across the type pools $\{\mathcal{V}_o\}$; within each type, representative nodes are sampled and summarized into a shared type context used during representation construction. The sampling distribution is then updated according to observed learning signals. In this sense, TypeBandit should be interpreted as a type-level resource-allocation bandit layered on top of heterogeneous representation learning.

This formulation is naturally connected to adversarial bandit algorithms such as Exp-style updates. In the implementation studied in this paper, TypeBandit maintains a nonzero exploration probability so that all candidate types retain a minimum chance of being selected. A typical smoothed type-level distribution has the form
\[
p_t(o)
=
(1-Kp_{\min})\frac{w_t(o)}{\sum_{o'\in\mathcal{A}}w_t(o')}
+
p_{\min},
\]
where $K=|\mathcal{A}|$, $w_t(o)$ is the current weight of type $o$, and $p_{\min}$ controls the minimum exploration probability. This type-level formulation keeps the adaptive module lightweight, since the policy operates over the small set of node types rather than over all nodes or edges. The choice is deliberate: a local bandit over every typed neighborhood would be far more expensive, while the present type-level policy still decides where limited sampling resources are spent.

The purpose of the bandit component is not to replace heterogeneous GNN backbones, but to provide a model-agnostic mechanism for allocating additional sampled context and reweighting node-type contributions according to their observed usefulness during attribute completion and representation learning. Importantly, the policy governs whether extra sampled type context is computed on top of a persistent base representation; it is therefore not a strict local partial-feedback bandit in which unsampled arms become completely unobserved. This makes TypeBandit compatible with multiple heterogeneous encoders while directly addressing type-dependent information asymmetry.

\section{Problem Definition}
\label{sec:pd}

Using the notation introduced in Section~\ref{sec:ps}, we study heterogeneous attribute completion under partial node-type observability.
\[
\mathcal{G}=(\mathcal{V},\mathcal{E},\mathcal{O}_{\mathcal{V}},\mathcal{R}_{\mathcal{E}})
\]
with node-type mapping $\varphi$ and edge-type mapping $\phi$. Only node types in $\mathcal{O}_{\mathcal{V}}^{+}$ have observed features, while node types in $\mathcal{O}_{\mathcal{V}}^{-}$ are attribute-missing. The observed feature matrices are
\[
\{\mathbf{X}_{o}\}_{o\in\mathcal{O}_{\mathcal{V}}^{+}},
\]
where $\mathbf{X}_{o}\in\mathbb{R}^{|\mathcal{V}_{o}|\times d_o}$. No raw feature matrix is available for $o\in\mathcal{O}_{\mathcal{V}}^{-}$.

The goal is to learn completed latent feature representations
\[
\mathbf{Z}^{(0)}=\{\mathbf{Z}^{(0)}_{o}\}_{o\in\mathcal{O}_{\mathcal{V}}},
\]
where each $\mathbf{Z}^{(0)}_{o}\in\mathbb{R}^{|\mathcal{V}_{o}|\times d}$ lies in a shared latent space. For attributed types, $\mathbf{Z}^{(0)}_{o}$ should preserve useful information from the observed attributes. For attribute-missing types, $\mathbf{Z}^{(0)}_{o}$ should be inferred from graph topology and cross-type information. These completed latent features are then passed to a heterogeneous graph backbone to produce final node embeddings
\[
\mathbf{H}=\{\mathbf{h}_{v}\mid v\in\mathcal{V}\}.
\]
The learned embeddings are evaluated through downstream tasks such as node classification.

A central difficulty is that the information available for completing missing attributes is not uniformly distributed across node types. Given a target node $v$, different typed neighborhoods $N_o(v)$ may provide substantially different levels of semantic relevance. In the implementation studied here, however, the adaptive decision is lifted from node-local neighborhoods to the coarser level of global type pools $\mathcal{V}_o$. For example, in an academic graph, paper nodes may provide strong textual or topical evidence for author representations, while other node types may provide weaker or more indirect signals. Conversely, in another dataset or under another backbone, the most useful type may differ. Therefore, a fixed aggregation or sampling strategy can be suboptimal because it cannot adapt to dataset-specific and training-stage-specific type importance or decide where a limited sampling budget should be spent.

We refer to this phenomenon as \textit{type-dependent information asymmetry}. Informally, type-dependent information asymmetry occurs when the expected utility of information from one node type differs from that of another type for the purpose of attribute completion and downstream prediction. In this setting, the model must solve two related problems:
\begin{enumerate}
    \item how to initialize latent features for attribute-missing node types; and
    \item how to allocate a finite type-level sampling budget and reweight different node types during representation learning.
\end{enumerate}

TypeBandit addresses these problems by learning a type-aware adaptive policy jointly with heterogeneous representation learning. Formally, given the graph $\mathcal{G}$, the observed attributes $\{\mathbf{X}_{o}\}_{o\in\mathcal{O}_{\mathcal{V}}^{+}}$, and a downstream supervision signal on a target node type, the objective is to learn
\[
\mathbf{Z}^{(0)},\quad \mathbf{H},\quad \text{and}\quad \mathbf{p}_{t},
\]
where $\mathbf{p}_{t}$ is a type-level sampling or weighting distribution updated during training. The desired solution should produce effective completed representations for nodes in $\mathcal{V}^{-}$, preserve useful information from attributed types in $\mathcal{V}^{+}$, and improve or maintain downstream performance when integrated with heterogeneous GNN backbones. Because a fully local bandit over every target-node neighborhood would be computationally prohibitive at scale, the formulation studied here instead uses a compact type-level resource-allocation policy over global type pools. A low-weight type may therefore receive few or even zero sampled nodes in a round, in which case the model falls back to its base latent representation for that type without adding extra sampled context.

Thus, the problem studied in this paper is adaptive heterogeneous attribute completion under type-dependent information asymmetry: learning latent features and node embeddings when observed attributes are available only for part of the heterogeneous graph, and when the usefulness of cross-type information varies across datasets, node types, and training stages. In our implementation, this takes the form of a type-level resource-allocation bandit that decides where additional sampled context should be computed rather than a node-local bandit that fully hides unsampled neighborhoods.

\section{Related Work}
\label{sec:work}

Related work on this problem can be grouped into three directions: heterogeneous graph representation learning, attribute completion for graphs with missing features, and adaptive sampling for graph neural networks.

\textbf{Heterogeneous graph representation learning.}
Heterogeneous graph neural networks have been widely studied for modeling graphs with multiple node and relation types. Representative backbone architectures include R-GCN~\cite{schlichtkrull2018modeling}, HetGNN~\cite{zhang2019heterogeneous}, HAN~\cite{wang2019heterogeneous}, MAGNN~\cite{fu2020magnn}, HGT~\cite{hu2020heterogeneous}, and SimpleHGN~\cite{lv2021simplehgn}. These methods differ in how they encode relation types, meta-path semantics, attention weights, and message passing mechanisms. More recent models such as SeHGNN~\cite{yang2023simple} and HINormer~\cite{mao2023hinormer} further improve efficiency and expressive power through simplified aggregation or transformer-style heterogeneous modeling. These architectures provide strong encoders for heterogeneous graph learning, but they generally assume that useful node features are already available or can be handled by standard preprocessing. Therefore, improved architecture design alone does not directly resolve the challenge considered in this work, where entire node types may suffer from missing or unreliable attributes.

\textbf{Attribute completion for graphs with missing features.}
A second line of work studies feature reconstruction or attribute completion on graphs. Topology-guided and propagation-based methods, such as FeatProp~\cite{rossi2022featprop}, use graph diffusion to propagate observed features to feature-missing nodes. Heterogeneous attribute-completion methods, including HetReGAT~\cite{li2023hetregat} and HeGAE~\cite{chen2024hegae}, further incorporate heterogeneous topology and higher-order neighborhood information to infer missing attributes. These methods are directly relevant to our setting because they address feature incompleteness rather than only downstream representation learning. However, most existing completion strategies are static once the propagation rule, meta-path, or reconstruction objective is specified. Although heterogeneous attention mechanisms can assign different weights to relations, meta-paths, or neighbors, these weights are usually optimized for downstream representation learning rather than for explicitly adapting type-level emphasis according to which node types are currently most helpful for attribute recovery. As a result, existing methods do not directly model type-level information asymmetry in the attribute-completion process. TypeBandit differs from this line of work by treating node-type informativeness as an adaptive signal learned through reconstruction-oriented feedback rather than as a fixed propagation design or an implicit attention weight.

\textbf{Adaptive sampling and bandit strategies for GNNs.}
Graph sampling methods have been extensively studied to improve the scalability and stability of GNN training. Representative approaches include FastGCN~\cite{chen2018fastgcn}, GraphSAINT~\cite{zeng2019graphsaint}, and Cluster-GCN~\cite{chiang2019cluster}, which reduce the cost of neighborhood aggregation through node-, edge-, subgraph-, or cluster-level sampling. Bandit-inspired sampling methods further adapt the sampling process during training. For example, EXP3-style and Exp4.P-style strategies have been used to adjust sampling probabilities according to observed rewards or approximation errors~\cite{auer2002nonstochastic,liu2020bandit,zhang2021biased,zhang2022hierarchical}. These works motivate our adaptive formulation, but their primary goal is efficient or low-variance neighborhood sampling for message passing. In contrast, TypeBandit shifts the role of the bandit mechanism from computational sampling efficiency to type-level resource allocation and signal reweighting for attribute completion. In our formulation, arms correspond to node types rather than individual nodes, edges, or generic sampling heuristics. The arm-selection process therefore operationally determines how much of a finite sampling budget is spent on each type, how many representative nodes are sampled from that type, and which additional global type contexts are computed for completion. The method is thus bandit-driven in its budget allocation and sampling decisions, but it remains deliberately type-level and global rather than a textbook node-local partial-feedback sampler.

\textbf{Stability, topology awareness, and positioning of TypeBandit.}
Recent analyses of graph-learning stability and topology-aware representation learning provide additional context for our design. Yang et al.~\cite{yang2025deeper} study stability and generalization in deep graph convolutional networks, reinforcing the importance of stable optimization when graph models become deeper or more iterative. Xie et al.~\cite{xie2026multitopology} emphasize representation learning across multiple graph topologies, which is complementary to our topology-aware and hybrid initialization design. In TypeBandit, topology-aware initialization provides a warm start for missing attributes, while a heuristic multiplicative-weights policy update subsequently adjusts type-level sampling weights during training. Thus, topology is used as an initialization signal rather than as a fixed completion rule. This combination distinguishes TypeBandit from both static attribute-completion methods and conventional sampling-based GNN acceleration techniques.

In summary, existing HGNN architectures provide increasingly strong heterogeneous encoders, and recent attribute-completion methods show that topology can be useful for reconstructing missing features. However, most prior methods rely on fixed propagation rules, predefined meta-paths, or architecture-level improvements, and therefore do not explicitly adapt to type-level information asymmetry. Existing bandit-based GNN samplers adapt neighborhood sampling, but they have not been developed for type-level heterogeneous attribute completion. TypeBandit addresses this gap by combining topology-aware initialization, heterogeneous representation learning, and adaptive type-level bandit sampling in a unified methodology.

\section{Methodology}
\label{sec:method}
TypeBandit is a front-end sampling, completion, and reweighting module for heterogeneous graph learning. Rather than assuming that every neighbor type is equally useful for attribute completion, it maintains a compact, learnable preference over node types and uses that preference to allocate a finite type-level sample budget and control how strongly each type contributes to the completed representation.

\begin{figure*}[t]
\centering
\resizebox{\textwidth}{!}{%
\begin{tikzpicture}[scale=0.75, transform shape,
    x=1cm,
    y=1cm,
    font=\small,
    >=Latex,
    box/.style={
        draw,
        rounded corners=3pt,
        align=center,
        inner sep=5pt,
        minimum height=1.1cm,
        text width=2.8cm
    },
    source/.style={box, fill=gray!8, draw=gray!60},
    init/.style={box, fill=blue!8, draw=blue!60!black},
    latent/.style={box, fill=teal!8, draw=teal!55!black},
    bandit/.style={box, fill=orange!14, draw=orange!75!black, text width=3.4cm},
    train/.style={box, fill=green!10, draw=green!50!black, text width=3.2cm},
    output/.style={box, fill=yellow!12, draw=yellow!45!black, text width=2.4cm},
    metric/.style={box, fill=gray!5, draw=gray!60, text width=2.6cm},
    group/.style={draw=gray!45, rounded corners=6pt, inner sep=14pt, dashed, thick},
    steplabel/.style={draw=gray!55, rounded corners=6pt, fill=gray!7, font=\small\bfseries, align=center, minimum width=2.2cm, minimum height=2.6cm},
    arrow/.style={->, thick, draw=gray!75, rounded corners=4pt},
    feedback/.style={->, dashed, thick, draw=orange!80!black, rounded corners=4pt},
    note/.style={draw=none, font=\scriptsize, align=center}
]

\node[steplabel] (s1) at (0.4, 8.2) {Step 1\\Initialization};
\node[steplabel] (s2) at (0.4, 4.8) {Step 2\\Type-aware\\construction};
\node[steplabel] (s3) at (0.4, -0.8) {Step 3\\Joint\\optimization};

\node[source] (attrs) at (4.2, 9.6) {Observed features\\$\{\mathbf{X}_{o}\}_{o\in\mathcal{O}_{\mathcal{V}}^{+}}$};
\node[source] (graph) at (4.2, 8.2) {Heterogeneous graph\\$\mathcal{G}$};
\node[source] (missing) at (4.2, 6.8) {Missing-feature nodes\\$\mathcal{V}^{-}$};

\node[init] (fprop) at (8.6, 9.6) {Feature propagation\\semantic prior};
\node[init] (degree) at (8.6, 6.8) {Relation-wise\\degree prior};
\node[latent] (proj) at (8.6, 4.8) {Latent projection\\shared feature space};
\node[train] (backbone) at (8.6, -0.8) {HGNN backbone encoder\\R-GCN / HetGNN /\\HGT / SimpleHGN};

\node[init, text width=3.2cm] (hybrid) at (13.0, 8.2) {Hybrid pretraining\\topology-aware warm start};
\node[latent] (gate) at (13.0, 4.8) {Gated fusion\\feature + topology};
\node[metric] (pred) at (13.0, 0.4) {Prediction head\\target logits};
\node[metric] (comp) at (13.0, -0.8) {Completion decoder\\masked features};
\node[metric] (embed) at (13.0, -2.0) {Target embeddings\\Macro-F1 / Micro-F1};

\node[bandit] (policy) at (17.4, 4.8) {TypeBandit policy\\sample budget + type weights};
\node[note] (policynote) at (17.4, 3.8) {adaptive type sampling and reweighting};
\node[output] (lpred) at (17.4, 0.4) {$\mathcal{L}_{prediction}$};
\node[output] (lcomp) at (17.4, -0.8) {$\mathcal{L}_{completion}$};
\node[metric] (reward) at (17.4, -2.2) {Reward proxy\\normalized type-wise norm};

\node[output, text width=1.6cm] (ltotal) at (21.6, -0.2) {$\mathcal{L}_{total}$};

\begin{scope}[on background layer]
    \node[group, fit=(graph)(attrs)(missing)(fprop)(degree)(hybrid)] (g1) {};
    \node[group, fit=(proj)(gate)(policy)(policynote)] (g2) {};
    \node[group, fit=(backbone)(pred)(comp)(embed)(lpred)(lcomp)(ltotal)(reward)] (g3) {};
\end{scope}

\draw[arrow] (attrs.east) -- (fprop.west);
\draw[arrow] (missing.east) -- (degree.west);

\coordinate (g_split_east) at ($(graph.east) + (0.7,0)$);
\draw[thick, draw=gray!75] (graph.east) -- (g_split_east);
\draw[arrow] (g_split_east) |- (fprop.195);
\draw[arrow] (g_split_east) |- (degree.165);

\draw[arrow] (fprop.east) -- ++(0.7,0) |- (hybrid.160);
\draw[arrow] (degree.east) -- ++(0.7,0) |- (hybrid.200);

\coordinate (a_split_west) at ($(attrs.west) - (0.6,0)$);
\draw[thick, draw=gray!75] (attrs.west) -- (a_split_west);
\draw[arrow] (a_split_west) |- (proj.west);

\draw[arrow] (hybrid.south) -- (gate.north);
\draw[arrow] (proj.east) -- (gate.west);
\draw[arrow] (gate.east) -- (policy.west);

\coordinate (g_split_west) at ($(graph.west) - (0.8,0)$);
\draw[thick, draw=gray!75] (graph.west) -- (g_split_west);
\draw[arrow] (g_split_west) |- (backbone.165);

\draw[arrow] (policy.south) -- ++(0,-1.8) -| (backbone.north);

\coordinate (b_split_east) at ($(backbone.east) + (0.6,0)$);
\draw[thick, draw=gray!75] (backbone.east) -- (b_split_east);
\draw[arrow] (b_split_east) |- (pred.west);
\draw[arrow] (b_split_east) |- (comp.west);
\draw[arrow] (b_split_east) |- (embed.west);

\draw[arrow] (backbone.south) -- ++(0,-2.0) -| (reward.south);

\draw[arrow] (pred.east) -- (lpred.west);
\draw[arrow] (comp.east) -- (lcomp.west);

\coordinate (lp_split) at ($(lpred.east) + (0.6,0)$);
\draw[arrow] (lpred.east) -- (lp_split) |- (ltotal.170);
\coordinate (lc_split) at ($(lcomp.east) + (0.6,0)$);
\draw[arrow] (lcomp.east) -- (lc_split) |- (ltotal.190);

\draw[feedback] (reward.east) .. controls +(6.0,0) and +(6.0,-2.0) .. (policy.east);

\end{tikzpicture}
}
\caption{TypeBandit Overview}
\label{fig:typebandit_methodology}
\end{figure*}
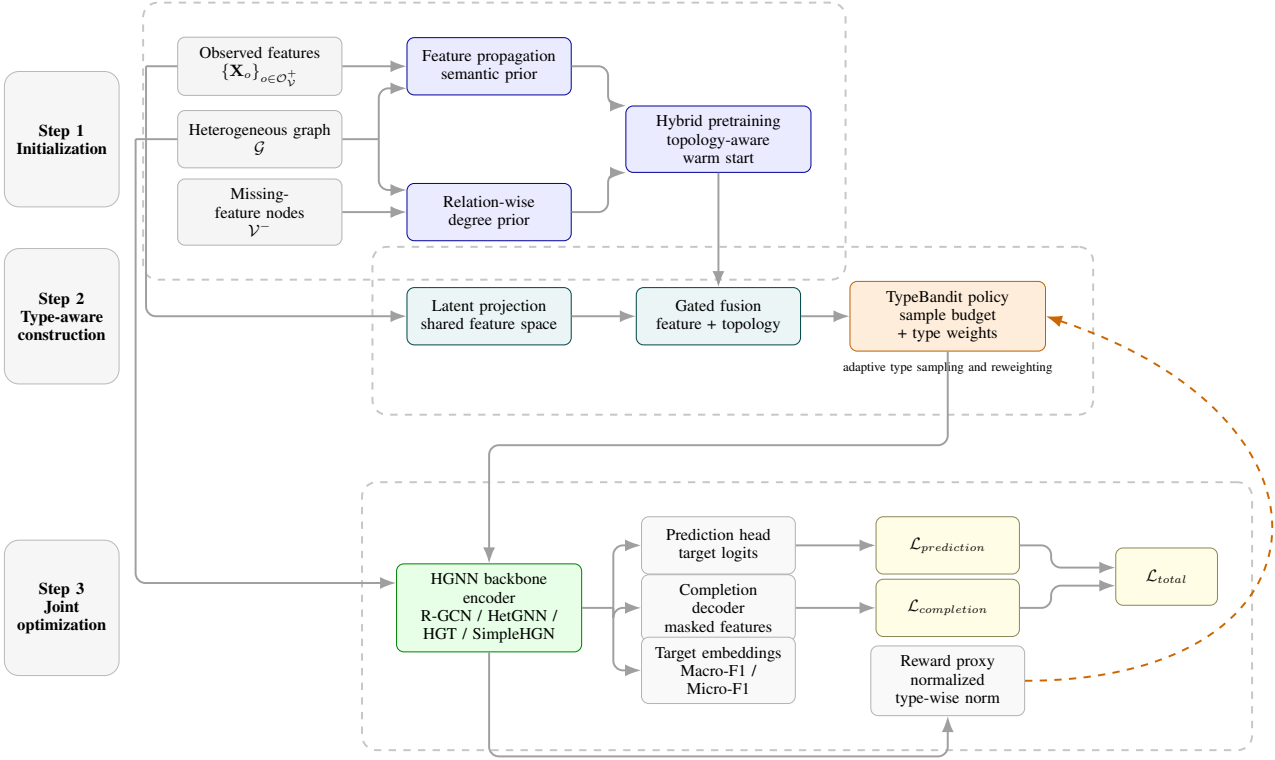

\subsection{Overview}

TypeBandit is designed to address adaptive heterogeneous attribute completion under type-dependent information asymmetry. Given a heterogeneous graph
\[
\mathcal{G}=(\mathcal{V},\mathcal{E},\mathcal{O}_{\mathcal{V}},\mathcal{R}_{\mathcal{E}}),
\]
only node types in $\mathcal{O}_{\mathcal{V}}^{+}$ are associated with observed attributes, while node types in $\mathcal{O}_{\mathcal{V}}^{-}$ have no raw feature matrix. The goal of TypeBandit is to construct completed latent feature matrices
\[
\mathbf{Z}^{(0)}
=
\{\mathbf{Z}^{(0)}_{o}\}_{o\in\mathcal{O}_{\mathcal{V}}},
\qquad
\mathbf{Z}^{(0)}_{o}\in\mathbb{R}^{|\mathcal{V}_{o}|\times d},
\]
and to adaptively control how different node types contribute to heterogeneous representation learning.

The implementation studied in this paper contains four interacting components. First, a topology-aware initializer produces structural priors for all node types, including the attribute-missing types in $\mathcal{O}_{\mathcal{V}}^{-}$. Second, observed attributes are projected into a shared latent space, while attribute-missing types receive a warm-start feature prior. Third, a type-level bandit policy maintains a probability distribution over candidate node types and converts this distribution into adaptive sampling budgets and type-level scaling factors. Fourth, a downstream heterogeneous backbone consumes the fused latent representations and is trained jointly with a downstream classification objective and a masked completion objective.

Let $\mathcal{A}\subseteq\mathcal{O}_{\mathcal{V}}$ denote the candidate type-level action space used by the bandit policy. Unless otherwise specified, we set $\mathcal{A}=\mathcal{O}_{\mathcal{V}}$ and $K=|\mathcal{A}|$. At training step $t$, TypeBandit maintains a type-level policy
\[
\mathbf{p}_{t}\in\Delta^{K-1},
\]
where $p_t(o)$ denotes the probability assigned to node type $o\in\mathcal{A}$. This policy determines how the finite context budget is allocated across node types and how strongly each type is reweighted before being passed to the heterogeneous encoder.

As illustrated in Figure~\ref{fig:typebandit_methodology}, the workflow proceeds from topology-aware initialization to latent feature construction, then to adaptive type-level sampling and reweighting, and finally to joint optimization. The figure separates the prediction branch, the completion branch, the $\lambda$-weighted loss combination, and the reward-feedback path that returns an embedding-derived reward proxy to the type-level policy. The implementation also supports an optional $H$-hop semantic-propagation prior, which we study later as an extension motivated by strong semantic encoders such as SeHGNN and by semantic-dominated regimes such as IMDB.

After TypeBandit constructs the completed latent inputs, a heterogeneous encoder $f_{\theta}$ produces final node embeddings:
\begin{equation}
\mathbf{H}
=
f_{\theta}\left(
\mathcal{G},
\{\mathbf{Z}^{(0)}_{o}\}_{o\in\mathcal{O}_{\mathcal{V}}}
\right).
\end{equation}
For a target node type $o_{\mathrm{tar}}$, the downstream prediction is obtained by
\begin{equation}
\hat{\mathbf{Y}}
=
\mathbf{H}_{o_{\mathrm{tar}}}\mathbf{W}_{\mathrm{cls}}
+
\mathbf{b}_{\mathrm{cls}}.
\end{equation}
The overall training objective combines the downstream prediction loss and the completion loss:
\begin{equation}
\mathcal{L}_{\mathrm{total}}
=
\mathcal{L}_{\mathrm{prediction}}
+
\lambda
\mathcal{L}_{\mathrm{completion}},
\end{equation}
where $\lambda\ge 0$ controls the contribution of the completion objective. Through this design, TypeBandit jointly learns completed latent representations, downstream node embeddings, and an adaptive type-level policy that reflects the changing usefulness of different node types during training.
\subsection{Topology-aware Initialization}

The first stage of TypeBandit constructs a topology-aware prior for every node type, including the attribute-missing types in $\mathcal{O}_{\mathcal{V}}^{-}$. This prior provides an initial structural representation before the bandit policy begins to adaptively reweight node types. In the present implementation, this stage supports three variants: degree-only pretraining, feature-propagation pretraining, and hybrid pretraining.

The degree-only variant is computed from relation-wise degree statistics across the heterogeneous schema. Let $\mathcal{R}_{\mathcal{E}}=\{r_1,\dots,r_m\}$ denote the set of relation types. For each node $v\in\mathcal{V}$, we construct a structural descriptor from its relation-specific degree profile:
\begin{equation}
\mathbf{s}_v =
\mathrm{Normalize}\Big(
\log(1+\mathrm{deg}_{r_1}(v)),
\ldots,
\log(1+\mathrm{deg}_{r_m}(v))
\Big),
\end{equation}
where $\mathrm{deg}_{r_i}(v)$ denotes the degree of node $v$ under relation type $r_i$, and $\mathrm{Normalize}(\cdot)$ denotes the same feature-wise normalization operator in both equations below. The topology embedding is then obtained through a type-specific projection:
\begin{equation}
\mathbf{t}^{\mathrm{deg}}_v =
\mathrm{Normalize}\left(
\mathbf{s}_v
\mathbf{W}_{\mathrm{topo}}^{(\varphi(v))}
+
\mathbf{b}_{\mathrm{topo}}^{(\varphi(v))}
\right),
\end{equation}
where $\mathbf{W}_{\mathrm{topo}}^{(\varphi(v))}$ and $\mathbf{b}_{\mathrm{topo}}^{(\varphi(v))}$ are the topology-projection parameters associated with the node type of $v$.

The feature-propagation variant uses observed attributes as anchors for topology-aware feature diffusion. Let $\tilde{\mathbf{x}}_v$ denote the projected observed feature of node $v$ when available, let $\mathbf{M}_{\mathrm{obs}}$ be the diagonal mask that is $1$ on attributed nodes and $0$ otherwise, and let $\bar{\mathbf{A}}=\sum_{r\in\mathcal{R}_{\mathcal{E}}}\mathbf{A}_r$ be the relation-collapsed adjacency with normalized operator $\bar{\mathbf{S}}=\bar{\mathbf{D}}^{-1/2}\bar{\mathbf{A}}\bar{\mathbf{D}}^{-1/2}$. In the default non-spectral feature-propagation variant, missing-feature nodes are initialized with the mean projected attributed feature
\begin{equation}
\boldsymbol{\mu}
=
\frac{1}{|\mathcal{V}^{+}|}
\sum_{u\in\mathcal{V}^{+}}\tilde{\mathbf{x}}_u.
\end{equation}
We then form the propagation-state matrix $\mathbf{U}^{(0)}\in\mathbb{R}^{|\mathcal{V}|\times d}$ row-wise by
\begin{equation}
\mathbf{U}^{(0)}_v
=
\begin{cases}
\tilde{\mathbf{x}}_v, & v\in\mathcal{V}^{+},\\
\boldsymbol{\mu}, & v\in\mathcal{V}^{-},
\end{cases}
\end{equation}
and run $H$ anchored propagation steps:
\begin{multline}
\mathbf{U}^{(h+1)}
=
\mathbf{M}_{\mathrm{obs}}\mathbf{U}^{(0)}
+
(\mathbf{I}-\mathbf{M}_{\mathrm{obs}})\bar{\mathbf{S}}\mathbf{U}^{(h)},\\
h=0,\dots,H-1.
\end{multline}
Thus observed-feature nodes are hard anchors at every hop, whereas missing-feature nodes absorb propagated semantic signal. The resulting propagation embedding is $\mathbf{t}^{\mathrm{prop}}_v=\mathbf{U}^{(H)}_v$.

The hybrid variant combines the degree-based topology embedding and the feature-propagation embedding. Let $\mathbf{t}^{\mathrm{prop}}_v$ denote the embedding obtained from the anchored propagation above. The hybrid topology prior is defined as
\begin{equation}
\mathbf{t}_v =
\rho_{\varphi(v)}
\mathbf{t}^{\mathrm{deg}}_v
+
(1-\rho_{\varphi(v)})
\mathbf{t}^{\mathrm{prop}}_v,
\end{equation}
where $\rho_{\varphi(v)}\in[0,1]$ is a type-specific mixing coefficient. In practice, the hybrid initializer is consistently stronger than degree-only pretraining and serves as the default initialization strategy in the reported experiments.

This initialization stage is important for two reasons. First, it gives attribute-missing node types meaningful latent priors before message passing begins. Second, it breaks the bootstrap circularity of the bandit policy: the initial reward signals are computed only after the encoder has received topology-aware latent inputs, rather than from uninformative random features.

\subsection{Latent Feature Initialization and Gated Fusion}

After constructing topology-aware priors, TypeBandit initializes latent features for all node types in a shared representation space. For attributed node types, the latent features are obtained by projecting observed attributes into the shared dimension $d$. For attribute-missing node types, which have no raw feature matrix, TypeBandit uses a warm-start feature prior derived from the attributed portion of the graph.

For each attributed node type $o\in\mathcal{O}_{\mathcal{V}}^{+}$, the observed feature matrix
\[
\mathbf{X}_{o}\in\mathbb{R}^{|\mathcal{V}_{o}|\times d_o}
\]
is projected into the shared latent space by a type-specific linear transformation:
\begin{equation}
\tilde{\mathbf{X}}_{o}
=
\mathbf{X}_{o}\mathbf{W}_{\mathrm{proj}}^{(o)}
+
\mathbf{1}
\left(\mathbf{b}_{\mathrm{proj}}^{(o)}\right)^{\top},
\qquad
o\in\mathcal{O}_{\mathcal{V}}^{+},
\end{equation}
where $\mathbf{W}_{\mathrm{proj}}^{(o)}\in\mathbb{R}^{d_o\times d}$ and $\mathbf{b}_{\mathrm{proj}}^{(o)}\in\mathbb{R}^{d}$ are type-specific projection parameters. The use of type-specific projections allows TypeBandit to handle heterogeneous raw feature dimensions across node types.

For attribute-missing node types, no raw feature matrix is available. We therefore construct a warm-start representation by averaging the projected features of attributed nodes:
\begin{equation}
\bar{\mathbf{x}}
=
\frac{1}
{\sum_{o\in\mathcal{O}_{\mathcal{V}}^{+}}|\mathcal{V}_{o}|}
\sum_{o\in\mathcal{O}_{\mathcal{V}}^{+}}
\sum_{u\in\mathcal{V}_{o}}
\tilde{\mathbf{x}}_{u}.
\end{equation}
Then, for each node $v\in\mathcal{V}_{o}$ whose type $o$ belongs to $\mathcal{O}_{\mathcal{V}}^{-}$, we initialize
\begin{equation}
\tilde{\mathbf{x}}_{v}
=
\bar{\mathbf{x}},
\qquad
v\in\mathcal{V}_{o},\quad
o\in\mathcal{O}_{\mathcal{V}}^{-}.
\end{equation}
This warm start gives missing-attribute node types a non-random feature prior before topology-aware fusion and heterogeneous message passing.

TypeBandit then combines the projected feature prior $\tilde{\mathbf{x}}_{v}$ with the topology-aware embedding $\mathbf{t}_{v}$ obtained from the previous stage. Since the relative usefulness of feature-based and topology-based priors may differ across node types, we use a type-specific gate:
\begin{equation}
\mathbf{g}_{v}
=
\sigma\left(
\mathbf{W}_{g}^{(\varphi(v))}
[
\tilde{\mathbf{x}}_{v}
\,\|\,
\mathbf{t}_{v}
]
+
\mathbf{b}_{g}^{(\varphi(v))}
\right),
\end{equation}
where $[\cdot\|\cdot]$ denotes vector concatenation and $\sigma(\cdot)$ is the sigmoid function. The initialized hidden representation of node $v$ is then
\begin{equation}
\mathbf{h}^{(0)}_{v}
=
\mathbf{g}_{v}\odot\tilde{\mathbf{x}}_{v}
+
(1-\mathbf{g}_{v})\odot\mathbf{t}_{v}.
\end{equation}

Equivalently, for each node type $o\in\mathcal{O}_{\mathcal{V}}$, the node-level vectors $\{\mathbf{h}^{(0)}_{v}:v\in\mathcal{V}_{o}\}$ form the initialized latent feature matrix
\[
\mathbf{H}^{(0)}_{o}
\in
\mathbb{R}^{|\mathcal{V}_{o}|\times d}.
\]
This representation is subsequently refined by the type-level sampling and reweighting module, producing the completed latent feature matrix $\mathbf{Z}^{(0)}_{o}$ used by the downstream heterogeneous encoder.

\subsection{Type-based Node Sampling}
\label{sec:ans}
The central idea of TypeBandit is to maintain an adaptive policy over node types rather than over individual neighbors. Let $\mathbf{p}^{(t)} \in \Delta^{K-1}$ denote the policy over the $K$ node types at training step $t$. In the implementation studied here, this policy should be interpreted as a type-level resource-allocation bandit: it decides how a finite global sampling budget is distributed across node types, not which exact neighborhood every target node should inspect. In the default implementation, the policy is updated using a heuristic multiplicative-weights rule with minimum exploration probability $p_{\min}$, where $0 \le p_{\min} < 1/K$:
\begin{equation}
p_k^{(t)} =
(1-Kp_{\min})\frac{w_k^{(t)}}{\sum_j w_j^{(t)}} + p_{\min}.
\end{equation}
In our default implementation, the reward is embedding-derived rather than directly attention-derived. We first compute a raw type score from the current encoded representations, using the mean embedding norm of each type as a lightweight proxy for present usefulness:
\begin{equation}
\tilde{r}_k^{(t)} = \frac{1}{|\mathcal{V}_k|}\sum_{v \in \mathcal{V}_k} \left\|\mathbf{h}^{(L)}_v\right\|_2.
\end{equation}
Before the bandit update, these scores are normalized across types so that the policy depends on relative usefulness rather than raw representation scale:
\begin{equation}
\bar{r}_k^{(t)} = \frac{\tilde{r}_k^{(t)}}{\sum_j \tilde{r}_j^{(t)} + \varepsilon}.
\end{equation}
The weights are then updated every $T$ epochs according to the heuristic multiplicative-weights rule used in the implementation:
\begin{equation}
w_k^{(t+1)} = w_k^{(t)} \exp\!\left(
\eta
\cdot
\left[\bar{r}_k^{(t)} + \frac{1}{p_k^{(t)}}\right]
\right),
\end{equation}
where
\[
\eta = \frac{p_{\min}}{2}\sqrt{\frac{\log N}{K T_{\mathrm{tot}}}}
\]
is the default code-level step size. Because $\eta$ depends on $N$ in the current implementation, changing $N$ alters both the sample budget and the update scale. Here $N$ is the base per-type context budget, $T_{\mathrm{tot}}$ is the scheduled maximum number of policy-update rounds under the training budget, and the inverse-probability term acts as an exploration-preserving bonus for low-probability types. In the default setting, this means $T_{\mathrm{tot}}=300/5=60$, so early stopping truncates realized training but does not redefine the scheduled step-size constant. Because $\bar{r}_k^{(t)}$ is normalized across types while $1/p_k^{(t)}$ can remain comparatively large near a near-uniform policy, the current evidence supports interpreting the update as reward-modulated, mild, exploration-preserving type-level reweighting rather than as a textbook EXP3/EXP4 rule or a strong reward-dominated type selector. This distinction is important: TypeBandit uses a bandit-style update to control the allocation of additional sampled context, but it is not intended to claim full equivalence to a strict node-local partial-feedback bandit.

\begin{figure}[!t]
\centering
\resizebox{\columnwidth}{!}{%
\begin{tikzpicture}[
    x=1cm,
    y=1.1cm,
    font=\small,
    panel/.style={draw=gray!55, rounded corners=5pt, fill=gray!3},
    title/.style={font=\bfseries\small, anchor=west},
    label/.style={font=\scriptsize, anchor=west},
    note/.style={font=\scriptsize\itshape, anchor=west, text=gray!70!black},
    budget/.style={draw=gray!55, rounded corners=4pt, fill=white, align=center, font=\scriptsize, inner sep=3pt},
    legend/.style={font=\scriptsize, anchor=west},
    emptyA/.style={circle, draw=blue!70!black, fill=white, minimum size=3.6mm, inner sep=0pt},
    fillA/.style={circle, draw=blue!70!black, fill=blue!35, minimum size=3.6mm, inner sep=0pt},
    emptyB/.style={circle, draw=teal!70!black, fill=white, minimum size=3.6mm, inner sep=0pt},
    fillB/.style={circle, draw=teal!70!black, fill=teal!35, minimum size=3.6mm, inner sep=0pt},
    emptyC/.style={circle, draw=orange!80!black, fill=white, minimum size=3.6mm, inner sep=0pt},
    fillC/.style={circle, draw=orange!80!black, fill=orange!35, minimum size=3.6mm, inner sep=0pt}
]

\node[legend] at (0.00, 0.5) {Budget $=6$; Type A: dense/weak, Type C: sparse/useful.};
\node[fillA] at (7.3, 0.5) {}; 
\node[legend] at (7.5, 0.5) {sampled};
\node[emptyA] at (8.7, 0.5) {}; 
\node[legend] at (8.9, 0.5) {available};

\draw[panel] (0, 0.1) rectangle (9.6, -2.2);
\node[title] at (0.2, -0.3) {Random sampling};
\node[note] at (0.2, -1.9) {uniform split across types};
\node[budget] at (8.3, -1.0) {fixed uniform\\rule\\$B=(2, 2, 2)$};

\node[label] at (0.2, -0.7) {Type A};
\foreach \x in {0,1} {\node[fillA]  at ({1.8 + 0.45*\x}, -0.7) {};}
\foreach \x in {2,3,4,5} {\node[emptyA] at ({1.8 + 0.45*\x}, -0.7) {};}

\node[label] at (0.2, -1.1) {Type B};
\foreach \x in {0,1} {\node[fillB]  at ({1.8 + 0.45*\x}, -1.1) {};}
\foreach \x in {2,3} {\node[emptyB] at ({1.8 + 0.45*\x}, -1.1) {};}

\node[label] at (0.2, -1.5) {Type C};
\foreach \x in {0,1} {\node[fillC]  at ({1.8 + 0.45*\x}, -1.5) {};}
\foreach \x in {2} {\node[emptyC] at ({1.8 + 0.45*\x}, -1.5) {};}

\begin{scope}[yshift=-2.6cm]
\draw[panel] (0, 0.1) rectangle (9.6, -2.2);
\node[title] at (0.2, -0.3) {Heuristic sampling};
\node[note] at (0.2, -1.9) {proportional rule favors dense pools};
\node[budget] at (8.3, -1.0) {proportional\\to pool size\\$B=(3, 2, 1)$};

\node[label] at (0.2, -0.7) {Type A};
\foreach \x in {0,1,2} {\node[fillA]  at ({1.8 + 0.45*\x}, -0.7) {};}
\foreach \x in {3,4,5} {\node[emptyA] at ({1.8 + 0.45*\x}, -0.7) {};}

\node[label] at (0.2, -1.1) {Type B};
\foreach \x in {0,1} {\node[fillB]  at ({1.8 + 0.45*\x}, -1.1) {};}
\foreach \x in {2,3} {\node[emptyB] at ({1.8 + 0.45*\x}, -1.1) {};}

\node[label] at (0.2, -1.5) {Type C};
\foreach \x in {0} {\node[fillC]  at ({1.8 + 0.45*\x}, -1.5) {};}
\foreach \x in {1,2} {\node[emptyC] at ({1.8 + 0.45*\x}, -1.5) {};}
\end{scope}

\begin{scope}[yshift=-5.2cm]
\draw[panel] (0, 0.1) rectangle (9.6, -2.2);
\node[title] at (0.2, -0.3) {TypeBandit (Ours)};
\node[note] at (0.2, -1.9) {learned reweighting toward useful types};
\node[budget] at (8.3, -1.0) {learned type\\weights\\$B=(1, 2, 3)$};

\node[label] at (0.2, -0.7) {Type A};
\foreach \x in {0} {\node[fillA]  at ({1.8 + 0.45*\x}, -0.7) {};}
\foreach \x in {1,2,3,4,5} {\node[emptyA] at ({1.8 + 0.45*\x}, -0.7) {};}

\node[label] at (0.2, -1.1) {Type B};
\foreach \x in {0,1} {\node[fillB]  at ({1.8 + 0.45*\x}, -1.1) {};}
\foreach \x in {2,3} {\node[emptyB] at ({1.8 + 0.45*\x}, -1.1) {};}

\node[label] at (0.2, -1.5) {Type C};
\foreach \x in {0,1,2} {\node[fillC]  at ({1.8 + 0.45*\x}, -1.5) {};}
\end{scope}

\end{tikzpicture}%
}
\caption{Type-Level Allocation Strategies}
\label{fig:sampling_comparison}
\end{figure}
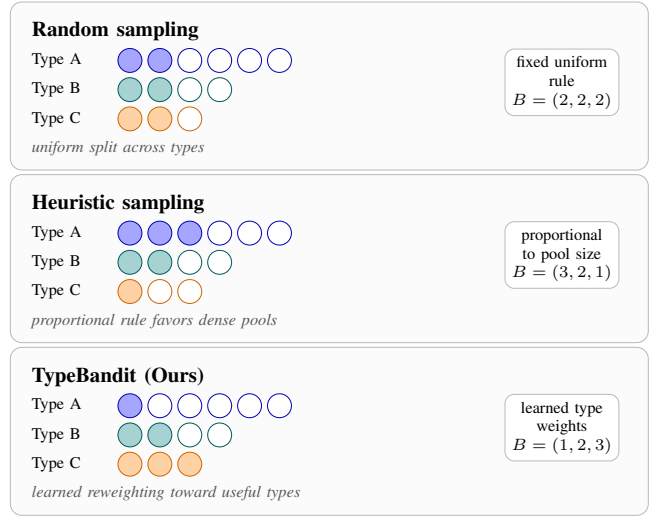

Figure~\ref{fig:sampling_comparison} contrasts random, heuristic, and adaptive type-level budget allocation on a toy heterogeneous graph.

In the actual forward pass, the learned policy first allocates a finite node-sampling budget across types. Because $N$ is the base per-type budget, the intended total context budget is approximately $KN$; after rounding and per-type clipping, type $k$ receives
\begin{equation}
B_k^{(t)} =
\min\!\left(|\mathcal{V}_k|,\max\!\left(0,\operatorname{round}(KNp_k^{(t)})\right)\right).
\end{equation}
Thus, $\sum_k B_k^{(t)}$ can differ slightly from $KN$ when rounding or clipping occurs, and low-weight types may receive zero sampled nodes in a given round. Within each type, TypeBandit samples a representative set $\mathcal{S}_k^{(t)} \subseteq \mathcal{V}_k$ with $|\mathcal{S}_k^{(t)}|=B_k^{(t)}$. The default within-type distribution is proportional to the current representation norm, while the uniform ablation uses uniform within-type sampling. When $B_k^{(t)}>0$, the sampled context is
\begin{equation}
\mathbf{c}_k^{(t)} = \frac{1}{|\mathcal{S}_k^{(t)}|}\sum_{u\in\mathcal{S}_k^{(t)}} \mathbf{h}^{(0)}_u .
\end{equation}
This context is fused back into every node of type $k$ through a lightweight gate:
\begin{multline}
\hat{\mathbf{h}}^{(0)}_v =
\mathbf{h}^{(0)}_v +
\beta_k \sigma\!\left(\mathbf{W}^{(k)}_s[\mathbf{h}^{(0)}_v \,\|\, \mathbf{c}_k^{(t)}]\right)
\\
\odot \left(\mathbf{c}_k^{(t)}-\mathbf{h}^{(0)}_v\right),
\qquad v\in\mathcal{V}_k.
\end{multline}
When $B_k^{(t)}=0$, no extra sampled context is constructed for type $k$ in that round and the model simply keeps $\hat{\mathbf{h}}^{(0)}_v=\mathbf{h}^{(0)}_v$ for $v\in\mathcal{V}_k$. Therefore, the bandit signal controls whether additional sampled type context is computed on top of the base latent state; it does not erase the base representation of an unsampled type.
Finally, the same policy is converted into a type-level scaling factor:
\begin{equation}
\alpha_k^{(t)} = 0.5 + p_k^{(t)},
\end{equation}
\begin{equation}
\mathbf{z}^{(0)}_v =
\mathrm{Dropout}\left(\hat{\mathbf{h}}^{(0)}_v \odot \boldsymbol{\gamma}_k \cdot \alpha_k^{(t)}\right),
\qquad v \in \mathcal{V}_k,
\end{equation}
where $\boldsymbol{\gamma}_k$ is a learnable type-specific scale vector and $\beta_k$ is a learned sampling-context strength. This design makes the sampling budget operational while keeping the policy interpretable and lightweight.

\subsection{Representation Learning}
\label{sec:rep}
After adaptive fusion and type-level reweighting, the completed latent inputs are passed to a downstream heterogeneous encoder. The final node embeddings and target-type logits are
\begin{equation}
\begin{aligned}
\mathbf{H}
&=
f_{\theta}\left(
\mathcal{G},
\{\mathbf{Z}^{(0)}_{o}\}_{o\in\mathcal{O}_{\mathcal{V}}}
\right),\\
\hat{\mathbf{Y}}
&=
\mathbf{H}_{o_{\mathrm{tar}}}\mathbf{W}_{\mathrm{cls}}
+
\mathbf{b}_{\mathrm{cls}}.
\end{aligned}
\end{equation}

To preserve information from the observed attribute space, TypeBandit also includes a lightweight feature decoder for each attributed type. The reconstruction term is
\begin{equation}
\mathcal{L}_{\mathrm{completion}} =
\frac{1}{|\mathcal{M}|}\sum_{v \in \mathcal{M}}
\left\|\mathbf{W}_{\mathrm{dec}}^{(\varphi(v))}\mathbf{h}^{(L)}_v + \mathbf{b}_{\mathrm{dec}}^{(\varphi(v))}
- \tilde{\mathbf{x}}_v\right\|_2^2,
\end{equation}
where $\mathcal{M}$ is the masked subset of attributed nodes used for reconstruction. The predictive term on the target nodes is standard cross-entropy:
\begin{equation}
\mathcal{L}_{\mathrm{prediction}} =
\frac{1}{|\mathcal{V}_{\mathrm{train}}|}\sum_{v \in \mathcal{V}_{\mathrm{train}}}
\mathrm{CE}(\hat{\mathbf{y}}_v, y_v).
\end{equation}
The total loss is
\begin{equation}
\mathcal{L}_{\mathrm{total}} = \mathcal{L}_{\mathrm{prediction}} + \lambda \mathcal{L}_{\mathrm{completion}},
\end{equation}
where $\lambda \ge 0$ is the completion-loss weight controlling the trade-off between supervised prediction and masked attribute reconstruction; setting $\lambda = 0$ removes the completion objective.

\subsection{Initialization, Convergence, and Algorithm Selection}
One natural concern about the reward-driven policy update is the bootstrap problem: if the policy update depends on signals derived from the current encoder state, how is the first meaningful reward obtained? TypeBandit resolves this circularity through a three-step initialization process. First, Stage~1 produces topology-aware priors that are independent of the bandit update itself. Second, the type-level bandit starts from uniform weights $w_k^{(0)} = 1/K$, so the first forward pass is unbiased across types. Third, only after this initial pass do we compute the first reward proxy from the encoded representations, which then bootstraps subsequent adaptive updates. In this way, topology-aware initialization and uniform exploration break the closed loop at the start of training.

Because the backbone parameters and bandit policy are updated jointly, the rewards are inherently non-stationary. Let $K$ be the number of node types, $T$ the number of policy updates, and
\begin{equation}
V_T = \sum_{t=1}^{T-1}\max_k |r_k^{(t+1)} - r_k^{(t)}|
\end{equation}
denote the total variation of the type-level rewards across training. Under the usual bounded-variation assumptions used in non-stationary bandit analysis~\cite{besbes2014stochastic}, regret terms of the form
\begin{equation}
R_T = O\!\left((V_T T^2 K \log K)^{1/3}\right).
\end{equation}
can arise. We do not present this expression as a formal theorem for the fully coupled TypeBandit-HGNN system; instead, we use it as theoretical intuition for why a multiplicative-weights-style update is a reasonable default when the reward distribution drifts as the heterogeneous encoder converges.

The choice of a multiplicative-weights-style update is also supported by algorithmic fit. Uniform allocation and $\epsilon$-greedy remain useful ablation baselines, but they do not directly target non-stationary reward adaptation. UCB-style rules, by contrast, are more naturally motivated by stationary reward assumptions. In our setting, the policy must respond to changing representation quality during training, which is better matched to an adversarial or non-stationary bandit view.

\begin{algorithm}[t]
\caption{TypeBandit Training Procedure}
\label{alg:typebandit_training}
\KwIn{Heterogeneous graph $\mathcal{G}$, observed features $\{\mathbf{X}_{o}\}_{o\in \mathcal{O}_{\mathcal{V}}^{+}}$, target labels $\mathbf{Y}$, max epochs $E_{\max}$, base sample budget $N$, update period $T$, exploration floor $p_{\min}$}
\KwOut{Trained encoder, final type-policy weights, target-type embeddings}
Construct topology-aware priors for all node types using Stage~1 pretraining\;
Initialize type-policy weights uniformly: $w_k \leftarrow 1/|\mathcal{A}|$ for each type $k$\;
\For{$epoch = 1$ \KwTo $E_{\max}$}{
Project observed attributes into the shared latent space and warm-start missing types\;
Fuse feature priors and topology priors through the type-specific gate\;
Allocate approximately $|\mathcal{A}|N$ context samples across types according to $\mathbf{p}^{(t)}$, subject to rounding and per-type clipping, and build sampled type contexts\;
Run the HGNN backbone to obtain target logits and decoder states\;
Compute $\mathcal{L}_{\mathrm{prediction}}$, $\mathcal{L}_{\mathrm{completion}}$, and $\mathcal{L}_{\mathrm{total}}$\;
Update model parameters by gradient descent\;
\If{$epoch \bmod T = 0$}{
Compute reward proxies from the current encoded representations\;
Update the multiplicative-weights type weights subject to the minimum exploration floor $p_{\min}$\;
}
\If{validation Macro-F1 has not improved for the patience window}{
stop early\;
}
}
\Return trained encoder, final type-policy weights, and target-type embeddings\;
\end{algorithm}

\subsection{Complexity}
The computational profile of TypeBandit is best understood by separating four cost layers rather than collapsing them into a single overhead term. First, the adaptive policy itself is compact: the policy state lives over $K=|\mathcal{A}|$ node types, so storing and updating the multiplicative-weights policy costs $O(K)$ per update round. Second, conditioned on a fixed initializer, the per-epoch front end consists of type-specific projections, gated fusion, and sampled type-context aggregation. The projection and gating terms scale as
\begin{equation}
O\!\left(\sum_{o \in \mathcal{O}_{\mathcal{V}}} |\mathcal{V}_{o}| d^2\right)=O(nd^2),
\end{equation}
because each node only passes through the parameters associated with its own type, while sampled type-context construction contributes an additional $O(\sum_{k=1}^{K} B_k d)$ for per-type sample budgets $B_k$.

Third, the Stage~1 topology-aware initializer may introduce one-time preprocessing costs that should be separated from the per-epoch front-end pass. Computing relation-wise degree descriptors requires a scan over typed edges, which is $O(|\mathcal{E}|)$ up to schema bookkeeping and can be written as $O(|\mathcal{E}| + n|\mathcal{R}_{\mathcal{E}}|)$ when the full relation-specific degree profile is materialized explicitly. Likewise, $H$-hop feature or semantic propagation contributes about $O(H|\mathcal{E}|d)$ preprocessing work. If a strengthened hybrid SVD regime is used, as in the IMDB extension study, the rank-$r$ truncated SVD introduces a further one-time spectral preprocessing cost determined by the dimensions of the propagated feature matrix and the chosen rank $r$.

Finally, typed message passing in the selected heterogeneous backbone still contributes the dominant graph-dependent training cost, and that term depends on whether the encoder is R-GCN, HetGNN, HGT, or SimpleHGN. This decomposition clarifies that only the adaptive policy state and update scale purely with $K$; the full TypeBandit front end still includes node-linear projection/gating work and, depending on the setting, one-time propagation or spectral preprocessing.

\section{Experiments}
\label{sec:exp}
We evaluate TypeBandit on three main heterogeneous graph benchmarks and complement them with an auxiliary sampled class-restricted OGBN-MAG academic benchmark. The empirical study focuses on six questions: (1) the predictive performance of the full model, (2) behavior across different HGNN backbones, (3) the contribution of completion and sampling modules, (4) the stability of the learned type-level policy, (5) computational overhead, and (6) scalability with respect to graph scale and schema complexity.

\subsection{Datasets and Evaluation Protocol}
We use DBLP, IMDB, and ACM with the benchmark instances used in our experiments. Only one node type in each dataset has observed attributes, while the remaining types are attribute-missing. The target node type differs across datasets: author for DBLP, movie for IMDB, and paper for ACM.

All main experiments follow a fixed-split protocol: DBLP uses 800/400/2857 train/validation/test nodes, IMDB uses 300/300/2339, and ACM uses 600/300/2125. We train on the provided masks, use early stopping on the validation split, and report test-set metrics averaged over 10 random seeds. Node classification is evaluated by Macro-F1 and Micro-F1 on the model outputs. Table~\ref{tab:dataset_stats} summarizes the heterogeneous graph statistics of DBLP, IMDB, ACM, and the full official OGBN-MAG graph for scale reference.

For the auxiliary OGBN-MAG analyses, we use sampled academic graphs rather than the full official graph. The larger inference-latency setting uses 100,000 target paper nodes and their associated heterogeneous context, while the class-restricted probe reported later uses 20k target paper nodes. Table~\ref{tab:dataset_stats} reports the full official OGBN-MAG graph scale for reference.

\begin{table*}[t]
\centering
\caption{Heterogeneous graph dataset statistics.}
\label{tab:dataset_stats}
\footnotesize
\setlength{\tabcolsep}{6pt}
\renewcommand{\arraystretch}{1.12}

\begin{tabular}{@{}l L{0.15\textwidth} r L{0.22\textwidth} r@{}}
\toprule
Dataset & Node Types & \# Nodes & Primary Relations & \# Edges \\
\midrule

DBLP
&
Paper (14,328) \newline
Author (4,057) \newline
Conference (20)
&
18,405
&
Paper--Author (19,645) \newline
Paper--Conference (14,328)
&
33,973 \\

\addlinespace[2pt]

IMDB
&
Movie (4,661) \newline
Actor (5,841) \newline
Director (2,270)
&
12,772
&
Movie--Actor (13,983) \newline
Movie--Director (4,661)
&
18,644 \\

\addlinespace[2pt]

ACM
&
Paper (3,025) \newline
Author (5,912) \newline
Subject (57)
&
8,994
&
Paper--Author (9,936) \newline
Paper--Subject (3,025)
&
12,961 \\

\midrule

OGBN-MAG
&
Paper (736,389) \newline
Author (1,134,649) \newline
Institution (8,740) \newline
Field (59,965)
&
1,939,743
&
Paper--Paper (5,416,271) \newline
Author--Paper (7,145,660) \newline
Author--Institution (1,043,998) \newline
Paper--Field (7,505,078)
&
21,111,007 \\

\bottomrule
\end{tabular}

\vspace{2pt}
\begin{minipage}{0.82\textwidth}
\end{minipage}
\end{table*}

\subsection{Backbones and Implementation Details}
We evaluate TypeBandit with four heterogeneous graph backbones: R-GCN, HetGNN, HGT, and SimpleHGN. Unless otherwise noted, all models use 64-dimensional hidden representations and are optimized with Adam using learning rate $5\times 10^{-3}$ and weight decay $5\times 10^{-4}$. We train with full-graph batches for at most 300 epochs, use early stopping with patience 50 on the validation split, set the completion-loss weight to $\lambda=0.4$, the minimum exploration floor to $p_{\min}=0.1$, and the propagation depth to $H=3$ for the feature- and semantic-propagation variants. The type-level policy is updated every 5 epochs.

The policy step size $\eta$ is not tuned independently. Following Section~\ref{sec:ans}, it is derived from $p_{\min}$, $N$, $K$, and $T_{\mathrm{tot}}$, where $T_{\mathrm{tot}}=60$ under the default 300-epoch schedule with policy updates every 5 epochs. Hybrid pretraining is used as the default Stage~1 initializer on DBLP and ACM. For IMDB, we additionally report a strengthened hybrid SVD plus semantic-propagation regime selected using the same fixed split and validation protocol; the test split is never used for model or regime selection.

All benchmark experiments were run on a single NVIDIA GeForce RTX 5080 GPU. Table~\ref{tab:hyperparams_current} summarizes the main optimization, architecture, and bandit hyperparameters used throughout the benchmark suite.

Broader matched comparisons against recent methods such as SeHGNN, HINormer, and FeatProp are left for future work, since robust codebase-aligned reruns of these methods were not available in the present experimental setup.

\begin{table}[!t]
\centering
\caption{Default Hyperparameters}
\label{tab:hyperparams_current}
\small
\resizebox{\columnwidth}{!}{%
\begin{tabular}{|l|l|c|}
\hline
Group & Hyperparameter & Value \\
\hline
Training & Optimizer & Adam \\
Training & Learning rate & $5\times 10^{-3}$ \\
Training & Weight decay & $5\times 10^{-4}$ \\
Training & Max epochs & 300 \\
Training & Patience & 50 \\
Training & Batch size & full graph \\
Initialization & Pretrain epochs & 50 \\
Initialization & Pretrain method & hybrid / hybrid SVD \\
Model & Hidden dimension & 64 \\
Model & Embedding dimension & 64 \\
Model & Num. layers & 2 \\
Model & Num. heads & 4 \\
Model & Dropout & 0.5 \\
Bandit & Completion weight $\lambda$ & 0.4 \\
Bandit & Exploration floor $p_{\min}$ & 0.1 \\
Bandit & Base sample budget $N$ & 20 \\
Bandit & Step size $\eta$ & $\frac{p_{\min}}{2}\sqrt{\frac{\log N}{K T_{\mathrm{tot}}}}$ (derived) \\
Initialization & Propagation hops $H$ & 3 \\
Bandit & Update period & 5 epochs \\
Bandit & Scheduled update rounds $T_{\mathrm{tot}}$ & 60 $(=300/5)$ \\
Bandit & Reward mode & norm \\
Protocol & Seeds & 10 \\
\hline
\end{tabular}
}
\end{table}

\subsection{Node Classification}
Table~\ref{tab:main_fixed} summarizes the main R-GCN results in the paper's primary evaluation setting. DBLP and ACM use the default hybrid initializer. IMDB is treated as an initialization-sensitive diagnostic benchmark: the reported R-GCN result uses a strengthened hybrid SVD plus semantic-propagation regime selected by validation performance only under the same fixed split and validation protocol because the default hybrid initializer is not the strongest regime on that dataset. Under these settings, TypeBandit reaches $92.53 \pm 0.60$ Macro-F1 on DBLP, $45.67 \pm 2.01$ on IMDB, and $82.46 \pm 1.30$ on ACM. The empirical pattern is therefore not one of uniform dominance: DBLP is the clearest positive case, ACM shows a moderate positive result, and IMDB is best read as a benchmark that reveals the method's dependence on initialization quality.
Unless a table is explicitly presented as a pretraining or extension study, this strengthened IMDB regime is the one used for the paper's main IMDB R-GCN row, including the matched backbone-only control, the regime summary, and the full ablation table.

Table~\ref{tab:backbone_only_rgcn_current} reports a matched stress-control comparison against a plain R-GCN backbone-only model under the same fixed-split protocol. We use this table to isolate the contribution of the TypeBandit front end under severe missing-feature conditions, not as the paper's main competitiveness comparison against stronger heterogeneous encoders. Under this control, TypeBandit improves over the plain backbone by $+6.60$ Macro-F1 on DBLP, $+4.14$ on IMDB, and $+1.70$ on ACM.

The DBLP stress-control margin still deserves interpretation rather than overstatement. In that benchmark, the target author nodes do not have observed attributes, so a plain R-GCN starts from weaker initial representations than the full TypeBandit model. We do not read this gap as evidence that all stronger heterogeneous encoders would fail similarly under matched preprocessing. For stronger same-setting reference points, Table~\ref{tab:backbones_current} is more informative: on DBLP, TypeBandit remains competitive across stronger backbones, with HGT and HetGNN slightly exceeding the default R-GCN configuration.

\begin{table}[!t]
\centering
\caption{Main Predictive Results with R-GCN}
\label{tab:main_fixed}
\small
\begin{tabular}{|l|c|c|}
\hline
Dataset & Macro-F1 & Micro-F1 \\
\hline
DBLP & $92.53 \pm 0.60$ & $93.33 \pm 0.56$ \\
IMDB & $45.67 \pm 2.01$ & $46.67 \pm 2.54$ \\
ACM & $82.46 \pm 1.30$ & $82.23 \pm 1.27$ \\
\hline
\end{tabular}
\end{table}

\begin{table}[!t]
\centering
\caption{Matched Stress Control with Plain R-GCN}
\label{tab:backbone_only_rgcn_current}
\small
\begin{tabular}{|l|c|c|c|}
\hline
Dataset & Backbone-only & TypeBandit & $\Delta$ \\
\hline
DBLP & $85.93 \pm 1.31$ & \best{92.53 $\pm$ 0.60} & $+6.60$ \\
IMDB & $41.53 \pm 1.25$ & \best{45.67 $\pm$ 2.01} & $+4.14$ \\
ACM & $80.76 \pm 0.69$ & \best{82.46 $\pm$ 1.30} & $+1.70$ \\
\hline
\end{tabular}
\end{table}

\subsection{Cross-backbone Evaluation}
Table~\ref{tab:backbones_current} compares TypeBandit across four heterogeneous encoders. The results show selective rather than uniform backbone compatibility. On DBLP, HGT and HetGNN slightly outperform R-GCN. On IMDB, the strengthened R-GCN configuration ($45.67 \pm 2.01$) and HGT ($45.59 \pm 2.23$) are statistically close, whereas HetGNN and SimpleHGN lag behind. The HGT row in this cross-backbone table is the harmonized benchmark setting; the separate $45.95 \pm 2.35$ HGT IMDB value reported later belongs to the dedicated 128-dimensional semantic-propagation extension study and should not be read as the same experimental regime. On ACM, R-GCN ($82.46 \pm 1.30$) and HGT ($82.75 \pm 0.95$) again form the strongest group, while HetGNN and SimpleHGN are notably weaker. SimpleHGN is substantially weaker on DBLP and less stable overall, indicating that backbone choice remains an important design factor.
We include SimpleHGN for completeness, but do not treat this untuned branch as the main evidence for compatibility.

\begin{table}[!t]
\centering
\caption{Cross-Backbone Results}
\label{tab:backbones_current}
\small
\begin{tabular}{|l|c|c|c|}
\hline
Backbone & DBLP & IMDB & ACM \\
\hline
R-GCN & $92.53 \pm 0.60$ & \best{45.67 $\pm$ 2.01} & $82.46 \pm 1.30$ \\
HetGNN & $93.72 \pm 0.45$ & $40.51 \pm 3.21$ & $66.86 \pm 5.71$ \\
HGT & \best{93.75 $\pm$ 0.30} & $45.59 \pm 2.23$ & \best{82.75 $\pm$ 0.95} \\
SimpleHGN & $24.32 \pm 1.08$ & $32.57 \pm 1.29$ & $67.81 \pm 10.79$ \\
\hline
\end{tabular}
\end{table}

\subsection{Ablation Study}
Table~\ref{tab:ablation_current} evaluates the contribution of the major components. The topology-only variant is clearly weaker than the full model on every dataset, which confirms that structural priors alone are insufficient without feature completion and downstream supervision. Removing pretraining hurts most on DBLP and especially on IMDB, while ACM is less sensitive. The remaining ablations reveal a more nuanced story than ``adaptive sampling always wins'': on IMDB, uniform, proportional, $\epsilon$-greedy, and context-removal variants all fall well below the full model, whereas on DBLP and ACM several simpler sampling variants are statistically similar to, and occasionally slightly above, the full run. Likewise, removing the completion module has little effect on DBLP and ACM but reduces IMDB from $45.67 \pm 2.01$ to $39.52 \pm 2.71$. We therefore interpret the gains as arising from the interaction among type-aware initialization, sampled type-context construction, completion supervision, and lightweight adaptive allocation. The adaptive policy is not intended to act as a universally aggressive selector; rather, it provides a stable type-level allocation mechanism whose benefit is most visible in harder, more asymmetric regimes such as IMDB.

To isolate the role of the Stage~1 initializer, Table~\ref{tab:pretraining_hybrid} compares four pre-SVD pretraining strategies on IMDB using the HGT backbone with 128-dimensional hidden representations. Degree-only pretraining is the weakest option, feature propagation closes most of the gap, and hybrid pretraining performs best among those non-spectral strategies, although its margin over the matched no-pretraining setting is small.

\begin{table}[!t]
\centering
\caption{Ablation Results}
\label{tab:ablation_current}
\small
\resizebox{\columnwidth}{!}{%
\begin{tabular}{|l|c|c|c|}
\hline
Variant & DBLP & IMDB & ACM \\
\hline
Full & $92.53 \pm 0.60$ & \best{45.67 $\pm$ 2.01} & $82.46 \pm 1.30$ \\
w/o pretrain & $91.24 \pm 0.38$ & $33.97 \pm 1.69$ & $82.07 \pm 0.92$ \\
Uniform sampling & $92.54 \pm 0.46$ & $39.10 \pm 2.26$ & \best{82.97 $\pm$ 0.65} \\
Proportional sampling & $92.52 \pm 0.50$ & $38.95 \pm 2.91$ & $82.57 \pm 0.77$ \\
w/o completion & $92.55 \pm 0.55$ & $39.52 \pm 2.71$ & $82.66 \pm 0.60$ \\
Topology-only & $67.75 \pm 1.13$ & $36.16 \pm 0.81$ & $61.27 \pm 1.10$ \\
$\epsilon$-greedy & \best{92.58 $\pm$ 0.50} & $39.02 \pm 2.83$ & $82.74 \pm 1.10$ \\
w/o policy scaling & $91.91 \pm 0.55$ & $37.14 \pm 3.74$ & $82.67 \pm 1.11$ \\
w/o sampling context & $92.54 \pm 0.64$ & $38.58 \pm 3.48$ & $82.86 \pm 1.28$ \\
\hline
\end{tabular}
}
\end{table}

\begin{table}[!t]
\centering
\caption{IMDB Pretraining Comparison with HGT}
\label{tab:pretraining_hybrid}
\small
\resizebox{\columnwidth}{!}{%
\begin{tabular}{|l|c|c|}
\hline
Pretraining & Macro-F1 & Micro-F1 \\
\hline
No pretraining & $44.48 \pm 1.57$ & $46.09 \pm 2.06$ \\
Degree-only & $37.51 \pm 4.01$ & $40.12 \pm 4.78$ \\
Feature propagation & $43.09 \pm 2.49$ & $44.40 \pm 2.97$ \\
Hybrid & \best{44.55 $\pm$ 2.18} & \best{46.24 $\pm$ 2.81} \\
\hline
\end{tabular}
}
\end{table}

\subsection{Stability Analysis}

Table~\ref{tab:stability_current} summarizes the stability of the learned type-level policy across 10 runs. The final arm-weight rankings are identical across all runs on DBLP, IMDB, and ACM, yielding a mean pairwise Kendall's $\tau$ of $1.0000$ on each dataset. This indicates that the learned type-level rankings are highly stable under repeated training.

Because the number of node types is small on the main benchmarks, Kendall's $\tau$ is used here only as a ranking-consistency diagnostic rather than as a high-dimensional stability metric. We therefore also report the mean final type weights and the minimum mean final weight to assess whether the policy collapses.

The final weights also remain far from a degenerate single-type allocation. The minimum mean final weight is $0.3252$ on DBLP, $0.3302$ on IMDB, and $0.3295$ on ACM, showing that the exploration-preserving update does not collapse the policy onto one node type. At the same time, the overall weight ranges are narrow, suggesting that TypeBandit performs mild reward-modulated reweighting rather than aggressive type filtering. This observation is consistent with the ablation results: on DBLP and ACM, where several simpler sampling variants perform similarly to the full model, the learned policy remains close to uniform; on IMDB, the full model benefits more clearly from the interaction between initialization, completion, and adaptive type-level context.

The top-ranked types are also interpretable under the corresponding graph schemas: conference receives the largest final weight on DBLP, director on IMDB, and subject on ACM.

\begin{table}[!t]
\centering
\caption{Type-Level Policy Stability}
\label{tab:stability_current}
\resizebox{\columnwidth}{!}{
\begin{tabular}{@{}l l c c c@{}}
\toprule
Dataset & Type & Mean final weight & Top type & Mean pairwise $\tau$ \\
\midrule
DBLP & paper      & 0.3252 & conference & 1.0000 \\
     & author     & 0.3281 &            &        \\
     & conference & 0.3467 &            &        \\
\midrule
IMDB & movie      & 0.3302 & director   & 1.0000 \\
     & director   & 0.3370 &            &        \\
     & actor      & 0.3328 &            &        \\
\midrule
ACM  & paper      & 0.3295 & subject    & 1.0000 \\
     & author     & 0.3301 &            &        \\
     & subject    & 0.3405 &            &        \\
\bottomrule
\end{tabular}
} 
\end{table}

\FloatBarrier
\subsection{Efficiency Analysis}
\begin{table*}[!t]
\centering
\caption{Training Efficiency Across Backbones}
\label{tab:efficiency_current}
\small
\begin{tabular}{|l|l|r|c|c|}
\hline
Dataset & Backbone & Params & Time (s/epoch) & Peak GPU (MB) \\
\hline
DBLP & R-GCN & 129490 & $0.0220 \pm 0.0024$ & 172.7 \\
DBLP & HetGNN & 146260 & $0.0195 \pm 0.0029$ & 175.1 \\
DBLP & HGT & 229970 & $0.0250 \pm 0.0019$ & 333.5 \\
DBLP & SimpleHGN & 104690 & $0.0256 \pm 0.0018$ & 190.3 \\
IMDB & R-GCN & 129360 & $0.0210 \pm 0.0030$ & 130.2 \\
IMDB & HetGNN & 146130 & $0.0188 \pm 0.0031$ & 134.6 \\
IMDB & HGT & 229840 & $0.0254 \pm 0.0038$ & 228.8 \\
IMDB & SimpleHGN & 104560 & $0.0246 \pm 0.0030$ & 142.2 \\
ACM & R-GCN & 129360 & $0.0211 \pm 0.0025$ & 101.7 \\
ACM & HetGNN & 146130 & $0.0175 \pm 0.0017$ & 106.0 \\
ACM & HGT & 229840 & $0.0234 \pm 0.0020$ & 172.5 \\
ACM & SimpleHGN & 104560 & $0.0236 \pm 0.0050$ & 111.9 \\
\hline
\end{tabular}
\end{table*}

\begin{table*}[!t]
\centering
\caption{Matched R-GCN Training Efficiency Control}
\label{tab:rgcn_efficiency_control}
\small
\begin{tabular}{|l|l|r|c|c|}
\hline
Dataset & Model & Params & Time (s/epoch) & Peak GPU (MB) \\
\hline
DBLP & Backbone-only & 33540 & 0.0140 & 94.4 \\
DBLP & TypeBandit & 129490 & 0.0220 & 172.7 \\
IMDB & Backbone-only & 33475 & 0.0137 & 75.5 \\
IMDB & TypeBandit & 129360 & 0.0210 & 130.2 \\
ACM & Backbone-only & 33475 & 0.0131 & 64.5 \\
ACM & TypeBandit & 129360 & 0.0211 & 101.7 \\
\hline
\end{tabular}
\end{table*}

\begin{table*}[t]
\centering
\caption{Matched Inference Latency Across Backbones}
\label{tab:matched_inference_latency}
\small
{\setlength{\tabcolsep}{6pt}
\renewcommand{\arraystretch}{1.12}
\begin{tabular}{llrrrr}
\toprule
Dataset & Backbone & Plain backbone (ms) & TypeBandit + backbone (ms) & Absolute overhead (ms) & Latency ratio \\
\midrule
IMDB & R-GCN & 7.99 & 15.36 & $+7.37$ & $1.92\times$ \\
IMDB & HGT & 6.86 & 12.83 & $+5.97$ & $1.87\times$ \\
\midrule
DBLP & R-GCN & 9.66 & 16.85 & $+7.19$ & $1.74\times$ \\
DBLP & HGT & 7.66 & 12.98 & $+5.32$ & $1.69\times$ \\
\midrule
ACM & R-GCN & 9.33 & 15.22 & $+5.89$ & $1.63\times$ \\
ACM & HGT & 5.94 & 12.58 & $+6.64$ & $2.12\times$ \\
\midrule
OGBN-MAG (100k) & R-GCN & 21.85 & 47.76 & $+25.91$ & $2.19\times$ \\
OGBN-MAG (100k) & HGT & 118.56 & 150.52 & $+31.96$ & $1.27\times$ \\
\bottomrule
\end{tabular}
}
\end{table*}
Table~\ref{tab:efficiency_current} summarizes the full-model training efficiency profile of TypeBandit across backbones, including parameter count, per-epoch training time, and peak GPU memory. HetGNN and R-GCN are typically the fastest options, whereas HGT is consistently the most memory-intensive and among the slowest; SimpleHGN is lighter in parameter count but does not provide a runtime advantage. To make the incremental training cost more concrete for the default backbone, Table~\ref{tab:rgcn_efficiency_control} reports a matched one-seed control against the plain R-GCN baseline. Under this comparison, TypeBandit uses about $3.9\times$ more parameters, takes roughly $1.5\times$--$1.6\times$ longer per epoch, and increases peak GPU memory by about $1.6\times$--$1.8\times$ across the three datasets. We therefore describe the additional training-time cost as real but still practical on the benchmark graphs considered here, rather than negligible.

In addition to training-time efficiency, we further report matched inference latency in Table~\ref{tab:matched_inference_latency}. This distinction is important because the bandit policy is updated during training but fixed during inference. Therefore, the inference-time comparison directly measures the deployment-time overhead of the TypeBandit-enhanced front end relative to the same plain backbone.

\textit{Matched inference latency.}
Table~\ref{tab:matched_inference_latency} reports a matched inference-latency comparison between plain HGNN backbones and TypeBandit-enhanced backbones. This comparison is designed to quantify the deployment-time overhead of adding the TypeBandit front end to the same backbone family. Since the type policy is learned during training and fixed during inference, no additional bandit update is performed at inference time.

All latency measurements use the same timing protocol. Here, the sampled OGBN-MAG probe uses \texttt{target\_limit} $=100{,}000$, and each latency ratio is computed relative to the corresponding plain backbone on the same dataset.

The results show that TypeBandit introduces measurable but practical inference overhead. On the three main benchmark datasets, the absolute overhead ranges from $5.32$ ms to $7.37$ ms for most settings, with ACM-HGT reaching $+6.64$ ms and a latency ratio of $2.12\times$ because the plain HGT backbone is particularly fast on this small graph. On the larger sampled OGBN-MAG probe, TypeBandit+R-GCN requires $47.76$ ms, adding $25.91$ ms over plain R-GCN. With the more expressive HGT backbone, plain HGT requires $118.56$ ms and TypeBandit+HGT requires $150.52$ ms, corresponding to a smaller relative latency ratio of $1.27\times$.

This pattern suggests that the relative overhead of TypeBandit depends on the computational cost of the chosen backbone. When the backbone computation is lightweight, the front-end cost accounts for a larger fraction of total latency; when the backbone is more expensive, as with HGT on sampled OGBN-MAG, the same TypeBandit-enhanced configuration contributes a smaller relative increase. We therefore interpret the deployment-time cost as nonzero but manageable in absolute latency, while the dominant scalability behavior remains tied to the selected HGNN backbone and graph-scale message passing. Overall, TypeBandit introduces measurable but manageable inference overhead, and its relative latency impact becomes smaller when the underlying HGNN backbone is more computationally expensive.

\subsection{Scalability Discussion}
The scalability of TypeBandit is governed primarily by the backbone encoder rather than by the type-level policy itself. The additional front-end cost consists of shallow projections, gated fusion, and a policy over $K$ node types, yielding an overhead that is linear in the number of nodes and modest in the number of types. In practical terms, the dominant cost remains heterogeneous message passing, while the TypeBandit-specific contribution is low-dimensional and grows gently with schema complexity.

This point is especially important for type diversity. The number of arms is the number of node types, not the number of nodes or relations, so the adaptive component scales with $K$ rather than with the full neighborhood size. On the benchmarks studied here, $K \in \{3,4\}$, and even in many real heterogeneous schemas the number of node types remains small enough that the bandit policy does not introduce a separate combinatorial bottleneck. The training-time and inference-time measurements in Tables~\ref{tab:efficiency_current}--\ref{tab:matched_inference_latency} are consistent with this view: runtime differences are dominated primarily by the backbone family and the scale of the evaluated graph setting.

This discussion remains conservative. The paper does not include a proprietary large-scale industrial benchmark, so we avoid making stronger empirical claims beyond the observed efficiency profile and the complexity analysis in Section~\ref{sec:method}. Nevertheless, both the theory and the measured runtimes indicate that TypeBandit should scale in step with the chosen backbone rather than become a standalone bottleneck. The key reason is structural: the adaptive policy operates over node types, so its state and update cost scale with $K$ rather than with the total number of nodes $n$. For industrial-style graphs with massive node counts but relatively small schema cardinality, this type-level decoupling is the main reason the adaptive component should remain manageable.

To probe this question further, we evaluate a sampled class-restricted OGBN-MAG academic benchmark. Table~\ref{tab:dataset_stats} reports the full official OGBN-MAG graph scale for reference, whereas Tables~\ref{tab:ogbn_mag_probe} and~\ref{tab:ogbn_mag_class_sweep} report class-restricted results obtained by sampling 20k paper nodes together with their directly connected academic context. This probe retains academic semantics, multiple node types, and a partially attributed setting, while still being large enough to stress the training setup. Under the R-GCN backbone, TypeBandit shows consistent positive gains across top-20, top-50, and top-100 class-restricted variants.

\begin{table*}[!t]
\centering
\caption{Sampled OGBN-MAG Probe Results}
\label{tab:ogbn_mag_probe}
\small
\begin{tabular}{|l|l|c|c|c|}
\hline
Backbone & Setting & Seeds & Macro-F1 & Micro-F1 \\
\hline
R-GCN & Top-50, backbone-only & 10 & $24.31 \pm 1.29$ & $37.39 \pm 1.25$ \\
R-GCN & Top-50, TypeBandit & 10 & \best{27.21 $\pm$ 1.56} & \best{39.04 $\pm$ 1.56} \\
HGT & Top-50, backbone-only & 3 & \best{25.73 $\pm$ 1.42} & \best{37.81 $\pm$ 2.49} \\
HGT & Top-50, TypeBandit & 3 & $24.33 \pm 1.05$ & $37.47 \pm 1.57$ \\
\hline
\end{tabular}
\end{table*}

\begin{table}[!t]
\centering
\caption{OGBN-MAG Class-Count Sweep}
\label{tab:ogbn_mag_class_sweep}
\small
\resizebox{\columnwidth}{!}{%
\begin{tabular}{|c|c|c|c|}
\hline
Top-$K$ classes & Backbone-only & TypeBandit & $\Delta$ Macro-F1 \\
\hline
20 & $38.68 \pm 0.51$ & \best{40.65 $\pm$ 1.18} & $+1.97$ \\
50 & $24.31 \pm 1.29$ & \best{27.21 $\pm$ 1.56} & $+2.90$ \\
100 & $13.73 \pm 1.17$ & \best{16.49 $\pm$ 0.45} & $+2.76$ \\
\hline
\end{tabular}
}
\end{table}

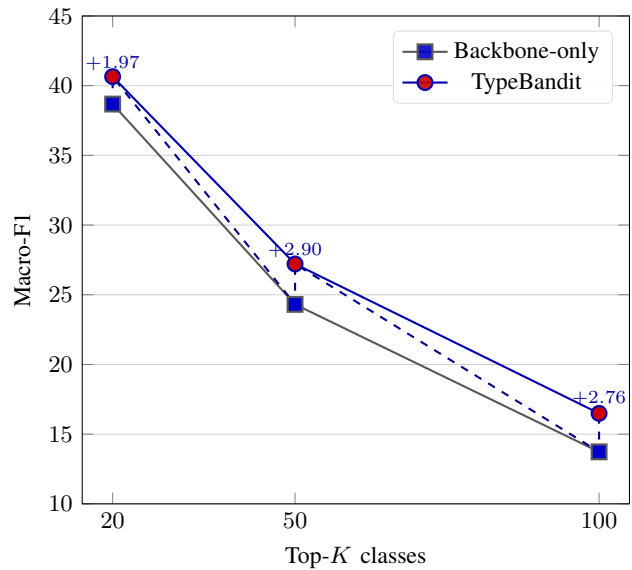
\begin{figure}[t]
\centering
\begin{tikzpicture}
\begin{axis}[
    width=\linewidth,
    height=0.33\textheight,
    xlabel={Top-$K$ classes},
    ylabel={Macro-F1},
    xmin=15,
    xmax=105,
    ymin=10,
    ymax=45,
    xtick={20,50,100},
    ymajorgrids,
    grid style={draw=gray!20},
    major grid style={draw=gray!35},
    tick label style={font=\small},
    label style={font=\small},
    legend style={
        font=\small,
        draw=gray!25,
        fill=white,
        fill opacity=0.9,
        text opacity=1,
        rounded corners=2pt,
        at={(0.97,0.97)},
        anchor=north east
    },
    clip=false
]
\addplot+[
    color=gray!70!black,
    thick,
    mark=square*,
    mark size=2.8pt
] coordinates {
    (20, 38.68)
    (50, 24.31)
    (100, 13.73)
};
\addlegendentry{Backbone-only}

\addplot+[
    color=blue!70!black,
    thick,
    mark=*,
    mark size=2.8pt
] coordinates {
    (20, 40.65)
    (50, 27.21)
    (100, 16.49)
};
\addlegendentry{TypeBandit}

\addplot[
    color=blue!55!black,
    dashed,
    thick,
    mark=none
] coordinates {
    (20, 38.68)
    (20, 40.65)
    (50, 24.31)
    (50, 27.21)
    (100, 13.73)
    (100, 16.49)
};

\node[font=\scriptsize, text=blue!60!black] at (axis cs:20,41.6) {$+1.97$};
\node[font=\scriptsize, text=blue!60!black] at (axis cs:50,28.2) {$+2.90$};
\node[font=\scriptsize, text=blue!60!black] at (axis cs:100,17.6) {$+2.76$};
\end{axis}
\end{tikzpicture}
\caption{OGBN-MAG Class-Count Sweep}
\label{fig:ogbn_mag_class_sweep}
\end{figure}

We do not over-interpret the absolute accuracy here, since the OGBN-MAG probe is class restricted rather than a full official benchmark; however, it still provides useful evidence that the methodology can be extended beyond the three medium-scale academic benchmarks and that the type-level policy does not become the dominant bottleneck as graph scale grows. In this sense, the probe serves as a public larger-scale point of reference for the industrial-scaling question while remaining close to the academic semantics of DBLP. This pattern suggests that the strongest practical benefits of TypeBandit arise when heterogeneous structure interacts with usable semantic attribute signal.

\FloatBarrier
\subsection{Hyperparameter Sensitivity}
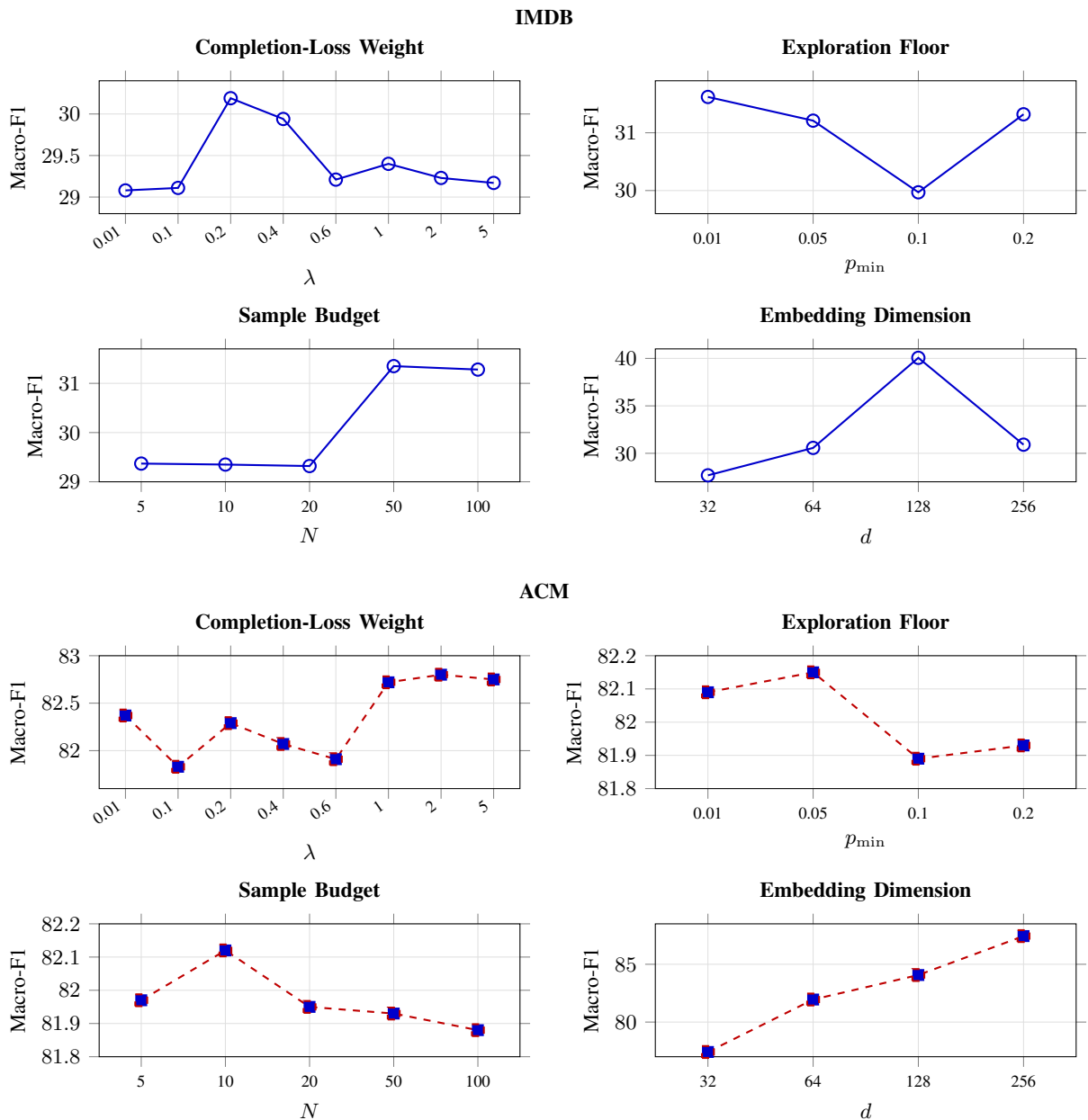
\begin{figure*}[!tb]
\centering

{\small\bfseries IMDB}\par\vspace{0.3em}
\begin{tikzpicture}
\begin{groupplot}[
group style={group size=2 by 2, horizontal sep=2.0cm, vertical sep=2.0cm},
width=0.43\linewidth,
height=0.145\textheight,
xlabel style={font=\small},
ylabel style={font=\small},
tick label style={font=\small},
title style={font=\small\bfseries},
ylabel={Macro-F1},
grid=major,
grid style={line width=.2pt, draw=gray!25},
tick align=outside,
clip=false,
]
\nextgroupplot[
title={Completion-Loss Weight},
xlabel={$\lambda$},
xmin=0.5,
xmax=8.5,
xtick={1,...,8},
xticklabels={0.01,0.1,0.2,0.4,0.6,1,2,5},
xticklabel style={font=\scriptsize, rotate=35, anchor=east},
ymin=28.8,
ymax=30.4,
]
\addplot+[mark=o, mark options={fill=white}, mark size=2.6pt, thick, color=blue!80!black] coordinates {(1, 29.080) (2, 29.110) (3, 30.190) (4, 29.940) (5, 29.210) (6, 29.400) (7, 29.230) (8, 29.170)};

\nextgroupplot[
title={Exploration Floor},
xlabel={$p_{\min}$},
xmin=0.5,
xmax=4.5,
xtick={1,...,4},
xticklabels={0.01,0.05,0.1,0.2},
xticklabel style={font=\scriptsize},
ymin=29.6,
ymax=31.9,
]
\addplot+[mark=o, mark options={fill=white}, mark size=2.6pt, thick, color=blue!80!black] coordinates {(1, 31.620) (2, 31.210) (3, 29.970) (4, 31.320)};

\nextgroupplot[
title={Sample Budget},
xlabel={$N$},
xmin=0.5,
xmax=5.5,
xtick={1,...,5},
xticklabels={5,10,20,50,100},
xticklabel style={font=\scriptsize},
ymin=29.0,
ymax=31.7,
]
\addplot+[mark=o, mark options={fill=white}, mark size=2.6pt, thick, color=blue!80!black] coordinates {(1, 29.370) (2, 29.350) (3, 29.320) (4, 31.350) (5, 31.280)};

\nextgroupplot[
title={Embedding Dimension},
xlabel={$d$},
xmin=0.5,
xmax=4.5,
xtick={1,...,4},
xticklabels={32,64,128,256},
xticklabel style={font=\scriptsize},
ymin=27.0,
ymax=41.0,
]
\addplot+[mark=o, mark options={fill=white}, mark size=2.6pt, thick, color=blue!80!black] coordinates {(1, 27.680) (2, 30.570) (3, 40.060) (4, 30.910)};
\end{groupplot}
\end{tikzpicture}

\vspace{0.8em}

{\small\bfseries ACM}\par\vspace{0.3em}
\begin{tikzpicture}
\begin{groupplot}[
group style={group size=2 by 2, horizontal sep=2.0cm, vertical sep=2.0cm},
width=0.43\linewidth,
height=0.145\textheight,
xlabel style={font=\small},
ylabel style={font=\small},
tick label style={font=\small},
title style={font=\small\bfseries},
ylabel={Macro-F1},
grid=major,
grid style={line width=.2pt, draw=gray!25},
tick align=outside,
clip=false,
]
\nextgroupplot[
title={Completion-Loss Weight},
xlabel={$\lambda$},
xmin=0.5,
xmax=8.5,
xtick={1,...,8},
xticklabels={0.01,0.1,0.2,0.4,0.6,1,2,5},
xticklabel style={font=\scriptsize, rotate=35, anchor=east},
ymin=81.6,
ymax=83.0,
]
\addplot+[mark=square*, mark size=2.5pt, thick, dashed, color=red!75!black] coordinates {(1, 82.370) (2, 81.830) (3, 82.290) (4, 82.070) (5, 81.910) (6, 82.720) (7, 82.800) (8, 82.750)};

\nextgroupplot[
title={Exploration Floor},
xlabel={$p_{\min}$},
xmin=0.5,
xmax=4.5,
xtick={1,...,4},
xticklabels={0.01,0.05,0.1,0.2},
xticklabel style={font=\scriptsize},
ymin=81.8,
ymax=82.2,
]
\addplot+[mark=square*, mark size=2.5pt, thick, dashed, color=red!75!black] coordinates {(1, 82.090) (2, 82.150) (3, 81.890) (4, 81.930)};

\nextgroupplot[
title={Sample Budget},
xlabel={$N$},
xmin=0.5,
xmax=5.5,
xtick={1,...,5},
xticklabels={5,10,20,50,100},
xticklabel style={font=\scriptsize},
ymin=81.8,
ymax=82.2,
]
\addplot+[mark=square*, mark size=2.5pt, thick, dashed, color=red!75!black] coordinates {(1, 81.970) (2, 82.120) (3, 81.950) (4, 81.930) (5, 81.880)};

\nextgroupplot[
title={Embedding Dimension},
xlabel={$d$},
xmin=0.5,
xmax=4.5,
xtick={1,...,4},
xticklabels={32,64,128,256},
xticklabel style={font=\scriptsize},
ymin=77.0,
ymax=88.5,
]
\addplot+[mark=square*, mark size=2.5pt, thick, dashed, color=red!75!black] coordinates {(1, 77.410) (2, 81.950) (3, 84.070) (4, 87.440)};
\end{groupplot}
\end{tikzpicture}

\caption{Hyperparameter Sensitivity on IMDB and ACM}
\label{fig:sensitivity_current}
\end{figure*}
Figure~\ref{fig:sensitivity_current} summarizes sensitivity trends for four representative hyperparameters: the completion-loss weight $\lambda$, the exploration floor $p_{\min}$, the base sample budget $N$, and the embedding dimension $d$. To keep the within-dataset trends visually interpretable, the figure separates IMDB and ACM into dataset-specific panels, uses equal-spaced horizontal positions for the tested hyperparameter values, and applies subplot-specific vertical ranges. We report these sweeps on IMDB and ACM because they represent the harder and moderate-gain regimes, respectively. On IMDB, the sweeps are shown for the default R-GCN setting rather than for the strengthened hybrid SVD plus semantic-propagation regime used for the best IMDB row in the main classification tables, so the plotted values should be read as within-setting sensitivity trends rather than as a reproduction of the strengthened IMDB peak. The $p_{\min}$ sweep directly addresses the exploration-parameter question within the valid range $p_{\min}<1/K$ and shows that performance remains relatively stable, with IMDB peaking around the more exploratory settings and ACM remaining nearly flat across the same range. Because $\eta$ depends on $N$ in the current implementation, the $N$ panel should be interpreted as a joint budget-and-step-size sweep rather than as a pure sample-budget ablation. We therefore treat those trends as reflecting both changes in sampled context and changes in policy-update scale. The additional panels show similarly moderate sensitivity to $\lambda$ and to this joint $N$/$\eta$ sweep, while the embedding-dimension sweep confirms that larger latent spaces can materially help on some benchmarks.

\subsection{Discussion}
The empirical picture that emerges from these experiments is nuanced and strongly dataset dependent. DBLP is the clearest positive case, ACM shows a smaller but positive gain regime, and IMDB is best understood as an initialization-sensitive diagnostic benchmark rather than as a universal-success case. Under the default IMDB setting the front end is not enough, whereas the validation-selected hybrid SVD plus semantic-propagation regime converts the matched R-GCN margin from negative to positive and lifts performance into the same range as HGT. Degree-only pretraining is consistently the weakest option in the focused initialization study, confirming that initialization quality matters in the harder settings. At the same time, the ablation study shows that the empirical gains are not attributable to initialization alone. Rather, they arise from the interaction among type-aware initialization, sampled type-context construction, completion supervision, and lightweight adaptive allocation. The adaptive policy is not intended to behave as a universally aggressive selector; instead, it provides a stable type-level allocation mechanism whose contribution is most visible when type-dependent asymmetry is stronger, as on IMDB, while on DBLP and ACM several simpler sampling variants remain statistically competitive with the full policy. The sampled OGBN-MAG academic probe sharpens this interpretation rather than overturning it: positive gains persist under R-GCN across top-20, top-50, and top-100 variants, but disappear under HGT. For large-scale or industrial-style graphs, the main takeaway is therefore not an absolute accuracy claim, but that the adaptive decision space remains type-level rather than node-level, so the extra control mechanism scales with schema size rather than graph size. Taken together, these findings position TypeBandit as a useful, dataset-dependent front end whose strongest benefits arise when heterogeneous structure carries meaningful type-specific signal and when initialization quality and type-level allocation work together rather than in isolation.

\FloatBarrier
\section{Conclusion}
\label{sec:con}
This paper studied heterogeneous graph attribute completion from the perspective of type-dependent information asymmetry and proposed TypeBandit as a type-aware adaptive completion methodology. The method combined topology-aware initialization, a type-level bandit policy that allocated finite sampling budget across node types, sampled type-context construction, and joint representation learning in a model-agnostic manner.

The experiments showed that TypeBandit was practical and stable under a fixed-split benchmark protocol on DBLP, IMDB, and ACM, but that its benefits were not uniform across datasets. The learned type-level policy was reproducible across seeds, the methodology was compatible with multiple HGNN backbones, and the computational overhead remained manageable. The clearest positive results appeared on DBLP, ACM showed a more moderate gain regime, and IMDB served primarily as an initialization-sensitive diagnostic benchmark. Reporting both its weaker default behavior and its stronger validation-selected hybrid SVD plus semantic-propagation regime clarified the boundary conditions under which TypeBandit was most effective without relying on test-set model selection. Beyond the main benchmarks, the sampled OGBN-MAG academic probe provided additional evidence of positive gains on a larger academic heterogeneous graph under simpler backbones such as R-GCN. For large-scale or industrial-style use, the key scaling argument is that TypeBandit's adaptive state depends on the number of node types rather than on the total number of nodes. Overall, the evidence supports TypeBandit as a useful, dataset-dependent type-aware front end under heterogeneous information asymmetry, not as a universal replacement for stronger semantic encoders or dataset-specific backbone tuning. A natural future direction is therefore to integrate the TypeBandit front end more tightly with stronger semantic encoders and to conduct matched comparisons against recent methods such as SeHGNN, HINormer, and FeatProp under the same fixed-split setting.

\section*{Acknowledgments}
This work has been supported by the U.S. National Science Foundation (NSF) under grant OAC-2209563 and CNS-2009057, as well as the DEVCOM Army Research Office (ARO) under grant W911NF2220159.

\bibliographystyle{IEEEtran}
\bibliography{tbd}

\vfill

\end{document}